\documentclass[conference]{IEEEtran}
\IEEEoverridecommandlockouts
\usepackage{cite}
\usepackage{graphicx}
\usepackage{amsmath,amssymb,amsfonts}
\usepackage{algorithmic}
\usepackage{textcomp}
\def\BibTeX{{\rm B\kern-.05em{\sc i\kern-.025em b}\kern-.08em
    T\kern-.1667em\lower.7ex\hbox{E}\kern-.125emX}}
\begin{document}


\newcommand{\jtext}[1]{{\color{blue} #1 \color{black}}}
\newcommand{\etext}[1]{#1}
\newcommand{\rev}[1]{{\color{red} #1 \color{black}}}
\renewcommand{\topfraction}{1.0}
\renewcommand{\bottomfraction}{1.0}
\renewcommand{\textfraction}{.01}
\renewcommand{\floatpagefraction}{1.0}
\renewcommand{\dbltopfraction}{1.0}
\renewcommand{\dblfloatpagefraction}{1.0}
\newcommand{\etal}{{et al. }}

\newcommand{\lw}[1]{\smash{\lower1.8ex\hbox{#1}}}
\newcommand{\emptybox}[1]{\begin{tabular}{|c|}\hline \\ ~\hspace{#1}~\\ ~\\ ~\\ ~\\ ~\\ ~\\ \hline \end{tabular}}

\newcommand{\tblcaption}[1]{\def\@captype{table}\caption{#1}}
\newcommand{\nakai}[1]{{\color{red} #1 \color{black}}} 
\newcommand{\onote}[1]{{\color{blue} #1 \color{black}}} 
\newcommand{\odel}[1]{{\color{blue} \sout{#1} \color{black}}} 
\newcommand{\rnote}[1]{}
\newcommand{\jtextd}[1]{} 
\newcommand{\etextd}[1]{}
\newcommand{\del}[1]{}

\def\Vec#1{\mbox{\boldmath $#1$}}

\newcommand{\figures}[1]{#1}

\newlength{\minitwocolumn}
\setlength{\minitwocolumn}{0.48\textwidth}
\newlength{\minithreecolumn}
\setlength{\minithreecolumn}{0.32\textwidth}
\newlength{\minifourcolumn}
\setlength{\minifourcolumn}{0.22\textwidth}

\newlength{\listlength}


\title{%
  Adversarial Example Generation using Evolutionary Multi-objective Optimization
%
}

\author{%
\IEEEauthorblockN{%
Takahiro Suzuki
}
\IEEEauthorblockA{%
\textit{Department of Information Science}\\
\textit{and Biomedical Engineering, }\\
\textit{Graduate School of Science}\\
\textit{and Engineering, }\\
\textit{Kagoshima University}\\
Kagoshima, Japan\\
sc115029@ibe.kagoshima-u.ac.jp}
\and
\IEEEauthorblockN{%
Shingo Takeshita
}
\IEEEauthorblockA{%
\textit{Department of Information Science}\\
\textit{and Biomedical Engineering, }\\
\textit{Graduate School of Science}\\
\textit{and Engineering, }\\
\textit{Kagoshima University}\\
Kagoshima, Japan\\
sc113035@ibe.kagoshima-u.ac.jp}
%
\and
\IEEEauthorblockN{%
Satoshi Ono}
\IEEEauthorblockA{%
\textit{Department of Information Science}\\
\textit{and Biomedical Engineering, }\\
\textit{Graduate School of Science}\\
\textit{and Engineering, }\\
\textit{Kagoshima University}\\
Kagoshima, Japan\\
ono@ibe.kagoshima-u.ac.jp}
%
}

\maketitle

\begin{abstract}
This paper proposes Evolutionary Multi-objective Optimization
(EMO)-based Adversarial Example (AE) design method that performs under
black-box setting.
Previous gradient-based methods produce AEs by changing all pixels of a
target image, while previous EC-based method changes small number of
pixels to produce AEs.
Thanks to EMO's property of population based-search, the proposed
method produces various types of AEs involving ones locating between AEs
generated by the previous two approaches, which helps to know the
characteristics of a target model or to know unknown attack patterns.
Experimental results showed the potential of 
the proposed method, e.g., it can generate
robust AEs and, with the aid of DCT-based perturbation pattern
generation, AEs for high resolution images.
%
\end{abstract}


\section{Introduction}

Over the past several years, deep learning has emerged as a ``go-to''
technique for classification.
In particular, object image recognition performance has been
significantly improved due to the rapid progress of Convolutional Neural
Networks (CNNs)~\cite{Krizhevsky2012}.
On the other hand, recent studies revealed that Neural Network
(NN)-based classifiers are susceptive to adversarial examples (AEs)\cite{Goodfellow2014,Chen2017,Ilyas2018,Nina2017,Papernot2017,Su2017,Athalye2018,Eykholt2018,Moosavi2017,Papernot2017,Guo2019,Shin2017}
that an attacker has intentionally designed to cause the model to make a
mistake.

AEs involve small changes $\Vec{\rho}$ to original images $\Vec{I}$ 
and fool the target NN as follows:
\begin{equation}
 \mathcal{C} \left( \Vec{I} + \Vec{\rho} \right) \neq 
  \mathcal{C} \left( \Vec{I} \right) 
\end{equation}
where $\mathcal{C}(\cdot)$ denotes classification result.
Such AEs can be easily generated using inside information of a target NN
such as gradient of loss function~\cite{Goodfellow2014}.
%

%
%

Considering practical aspects, there are many cases that the inner
information of target models cannot be available, e.g., commercial or
proprietary software and services.
Therefore, some studies attempted to attack NNs under black-box
setting where the attacker cannot access to the gradient of the
classifier~\cite{Chen2017,Ilyas2018, Nina2017,Papernot2017,Su2017}.
Under the black-box setting, Evolutionary Computation (EC) is expected
to play an important role.
In fact, one of the previous work~\cite{Su2017} employed Differential
Evolution (DE)~\cite{Storn1997}.
%
The previous work that directly uses EC changed one or a very small
number of pixels because this method must determine both which pixels
and how strong the pixels should be perturbed.
%
%
In opposite, methods under white-box setting such as the gradient-based method are likely to
change many pixels of a target image.
It is meaningful to comprehensively generate various AEs including ones
locating between AEs generated by EC and gradient-based method from the viewpoint of both
creating unknown kind of AEs and knowing the characteristics of NN
deeper.

\del{
robust:
\cite{Shin2017,Ilyas2018,Athalye2018}.

包括的にサンプルを作りたい

より現実的なのはblack-box approach．
ECによる生成が期待できる．
なお，DNNの構造自体を（進化型多目的最適化の基礎である）進化計算により設
計する手法[Real2017]が最難関国際会議ICMLに採録されたことからも，進化計算
の大域的最適化の能力が改めて注目されている．
}

By the way, generating AEs essentially involves more than one objective
function that have trade-off relationship such as classification
accuracy versus perturbation amount.
Most AE design methods put them together into single objective function
by linear combination, and, to the best of our knowledge, no study
attempted to generate AEs without integrating the objective functions in
a multi-objective optimization (MOO) manner.

Therefore, this study proposes an evolutionary multi-objective
optimization (EMO) approach for AE generation.
Thanks to population-based search characteristics of EMO, The proposed
method can generate AEs under black-box setting.
In addition, taking the advantages of population-based search of EMO,
the proposed method generates various AEs
such as robust AEs against image transformation.
Experimental results on representative datasets of CIFAR-10 and
ImageNet1000 have shown that the proposed method can generate various
AEs locating between the EC- and gradient based previous methods, and
attempt have been conducted to generate robust AEs against image
rotation.

%

\del{

本研究の目的は，機械学習により生成された学習器の特性の評価を目的として敵
対的サンプルを供出し，頑健な学習器の実現に向けた一助となることである．
\この目的を実現するために，進化型多目的最適化[Zitzler1999]を用いた敵対的
サンプルの網羅的生成方式を開発する．
進化型多目的最適化は，多点探索型のメタヒューリスティクスであり，トレード
オフ関係にある複数の目的関数を同時に最適化することでパレート解集合を導出
する．
解候補（敵対的サンプル）群の生成と評価を繰り返し，敵対的サンプルを網羅的
に生成することで，
学習器が持つ潜在的な脆弱性を暴露し，より頑健性の高い機
械学習モデルを設計するための知見を供出する．

}

We summarize the contributions of this paper as follows:
\begin{itemize}
 \item {\bf The first attempt to design AEs using EMO:}
	   which allows flexible design of objective functions and
	   constraints; non-differentiable, multimodal, noisy functions can
	   be used.
%


 \item {\bf Robust AE generation under black-box setting:}
	   Previous work designing robust AEs optimize expected
	   classification probability~\cite{Athalye2018,Eykholt2018}; however, considering only
	   averaged accuracy might generate AEs that can inappropriately be
	   classified its correct class in rare cases.
	   Taking the advantage of EMO, the proposed method supplementarily
	   employs its deviation as second objective function, allowing to
	   generate more robust AEs.

 \item {\bf DCT-based method:}
	   To generate AEs for high resolution images, the proposed method
	   designs perturbation patterns on frequency coefficients obtained
	   by two-dimensional Discrete Cosine Transform (2D-DCT)~\cite{Rao1990}, resulting
	   in reducing the dimension of the design variable space.
\end{itemize}

\section{Related Work}




The most popular approach to generate adversarial examples is to adopt
gradient of loss function in a target classifier under white-box
setting~\cite{Goodfellow2014}.
%
%
It generates AEs by simply adding the small perturbation to all pixels
of a target image according to gradient of a loss function.

Recently, universal perturbation that is applicable arbitrary images
and can lead NNs to make a misclassification~\cite{Moosavi2017}.
Interestingly, the perturbation pattern works well not only for the NN
used to design the pattern but also other NNs.
However, because it is a universal pattern, once the pattern is known, it
can be easily detected.

From the practical viewpoint, AE design methods that can work under
black-box setting are desirable;
such method allows to analyzing the characteristics of the consumer or
proprietary software or services.
In addition, different types of AEs from ones generated by
gradient-based methods help to further analysis of target NN models.
%
%
Su et al. proposed one pixel attack method~\cite{Su2017} using Differential
Evolution~\cite{Storn1997} that revealed the fragileness of the
classifiers.
Nina et al. proposed a local search method that approximates the
network gradient~\cite{Nina2017}. 
The above methods \cite{Su2017,Nina2017} changes small number pixels of
a target image to mimic the target classifier, whereas the
gradient-based method changes all pixels.
These two approaches produced different types of AEs.
Discovering various AEs is useful from both the viewpoint of knowing the
characteristics of NN more deeply and knowing an unknown attack
patterns.
That is the motivation we introduce multi-objective optimization for AE
design.

\del{

\begin{itemize}
 \item 勾配を利用する方法
 \item universalなAE

	   Universal AE ... universalなために，パターンの知識を知っている人は見破れ
	   る．

	   UniversalなAEも存在． 複数のclassifierに横断的に適用できる点が特にすごい．
	   一方で，パターンが既知となってしまうことは欠点．
	   ECを使えば，既存のUAEとは異なるパターンの生成も可能に．

 \item blackbox

	   [Papernot2016] 手元でかわりのモデル（substitute）に学習させてAEを
	   生成． あるモデルで作成したAEは他のモデルもだませることが多いため．
	   自分でtraining setを用意． oracleの呼び出し回数を限定？ （これは
	   提案手法において問題になりそう．）

	   NNに限定される

	   [Nina2017] ごく少数の画素を変更してgrdientをapproximate．

	   perturbationの与え方: sign-preserving perturbation function

	   目的関数: NNが割り当てる確率

	   直前のステップで変更した画素の周辺を変更
	   （直前のステップの変更は生きたまま）

	   translformation function $g$

	   パラメータ多い？: 画素数$t$，変更する画素の距離$d$，CYCLICに従っ
	   て変更する幅？$r \in [0, 2]$ ，変更する輝度値幅$p$（範囲制限なし，
	   自動的に決定），$LB$，$UB$

	   局所解に陥った場合の対策がなく，初期状態に依存して良好な解が得ら
	   れない危険がある．

	   [Nina2017]は少量の画素に対して摂動を加えることを繰り返すことでAE
	   を生成する方法を提案している． 
	   CIFAR10，STL10，SVHN，MNISTの4つのデータセットで数パーセントの画
	   素のみを変更する方法を実現している． ImageNet1000で0.5パーセント
	   程度の画素のみを変更．

	   [Chen2017]

	   []

	   FGSMは全画素を変更．

 \item robustなAE

	   道路標識などを対象に，視点が変化しても頑健に誤認識を誘発する
	   robustなAEの研究が行われている．
	   \cite{Eykholt2018} ... whitebox approach, 期待値を最小化（式(3))

	   \cite{Athalye2018} ... Stochastic gradient descentでtarget class
	   にmissclassifyする確率(の期待値)を最大化，LAB空間におけるL2ノルムを最小化す
	   る項を正則化項として考慮．
	   
	   JPEG圧縮に対してrobustなAEも提案\cite{Shin2017}されている．
	   gradientが必要．

	   blackbox ...ない

	   multi-objの強みの1つとして，robustなAEを作れる点を示す．

 \item DCTを利用するAE

	   DCTを使う方法...concurrentに研究されている[Guo2019]

\end{itemize}


\begin{itemize}
 \item 1画素だけ変更するAE

	   ECを使う方法...one pixel attack[Su2018]．画素が多いと難しい．

 \item 九大の手法
	   
\end{itemize}

}

\section{The Proposed Method}

\subsection{Key Idea}

{\bf 1. Formulating an adversarial pattern design problem as
multi-objective optimization: }
AE design problem essentially consists of more than one cost functions
that compete with each other such as accuracy versus visibility.
Therefore, it is natural to solve the problem without integrating them
into single objective function 
in accordance with the way of multi-objective optimization.
The proposed method does not require any parameters to integrate the
functions, 
and allows considering non-differentiable and/or non-convex objective
functions.
For instance, introducing two functions of the number of perturbed
pixels ($l-0$ norm) and the strength of the perturbation ($l-1$
norm) isolately allows clarifying the trade-off relationship between
them.
Decision makers can choose the most balanced AE from the Pareto optimal
solutions while considering target image properties.

\del{

AE生成の問題は本質的に，
正解クラスへ誤認識させる確率（最小化）とperturbationの小ささ（最小化）の
相反する2つの目的関数を持つ．
これまでの研究では，それらの線形和をとることで単一目的最適化にモデル化し
たり，片方を制約条件とするなどのアプローチをとっていた．
しかし，そのためにはパラメータや閾値を設定する必要があり，試行錯誤を必要
とする．
なお，FGSMや[Nina2017]などの先行研究と比べて，必ずしも優れた（すなわち少
ない画素数で）AEを生成できる訳ではないものの，より多くの観点で設計された
AEを包括的に生成できる．
これら2つの目的関数のパレート解は多数存在し，多様なAEを生成することは一
つの利点となる．

他の多くのAE生成方式と同様，特定のクラスに誤認識させることも可能
（targeted misclassification）

}

{\bf 2. Applying Evolutionary Multi-objective Optimization (EMO) algorithm:} 
The proposed method adopts an EMO algorithm to perform MOO.
Compared to the approach that trains substitute models\cite{Papernot2017},
the proposed method does not need to train the substitute model and is 
applicable models other than NNs.
In addition, thanks to EMO's essential property of population-based
search, 
the proposed method comprehensively produces non-dominated solutions.
%
Although there is no guarantee that the proposed method produces better
AEs than previous work,
finding various AEs with the proposed method helps 
to know the characteristics of a target NN model more deeply or to know
unknown attack patterns.

Furthermore, EMO does not require that the objective function be
differentiable, smooth, and unimodal, then various types of objective
functions and constraints can be used in the proposed method.
For instance, the proposed method can produce AEs more robust against
image transformation by adding standard deviation of classification
accuracy into objective functions in addition to the expected accuracy.

\del{

また，Universal AEは単一のperturbationで多様なモデルを騙せるものの，パター
ンが既知となると容易に防がれてしまう．
提案手法は個々の画像に特化したAEを生成することで，公に知らせていない未知
のAEを生成できる．

提案方式は互いにトレードオフの関係にある複数の指標に基づいてパターンの生
成を行うため，網羅的に敵対的サンプルを生成できる．
例えば，第１目的関数として摂動を加える画素の総数，第２目的関数として摂動
を加える輝度値幅の最大値の2つの指標を同時に最小化することにより，一部の
画素を強く変更するサンプル\cite{}から，FGSMのように多くの画素を微小に変
化するサンプル\cite{}までを同時に生成できる．A網羅的に生成されたサンプル
の設計変数および目的関数に着目した解析を行うことで，対象のモデルの脆弱性
に関する知見の導出が可能となる．

誤認識率の期待値のみを最大化する従来手法[Athalye2017]では，例外的に正し
く認識されてしまうサンプルの生成の抑制が困難である．
提案方式では，誤認識率の標準偏差を第2目的関数として同時に最小化[6]するこ
とで，安定的に誤認識を誘発する敵対的サンプルを生成できる．
逆に，ある特定の状況下でのみ誤認識を引き起こす敵対的サンプルの創出も可能
である．

}

{\bf 3. Black-box approach: } 
Taking one of the EMO's advantages, i.e., population-based search, the
proposed method performs under black-box
setting~\cite{Papernot2017,Nina2017,Chen2017},
which means that the proposed method does not require gradient information in a
target model; classification results involving assigned labels and
corresponding confidence are sufficient%
\footnote{%
Utilizing the high degree of freedom of the proposed method in the
design of the objective functions, even the confidence is unnecessary.
}.
Therefore, the proposed method is applicable to proprietary systems
and models other than NNs.

\del{
提案する方式は，ニューラルネットワーク内部
の情報を必要とせずにblack-box的にAEを生成する．
従来の勾配を必要とする方法は，プロプライエタリなサービスなどに応用できず，
実用性が高くなく，どちらかというと，NNのsensitivityを評価する手段ともい
える．
多くのAE生成方式が対象とするモデルの損失関数の勾配を必要とするのに対して，
提案方式はBlackbox的なアプローチをとり，モデルの内部情報を必要としない
%
black-boxアプローチは，認識結果のクラスとその確率のみを対象から受け取る
\cite{Papernot2017,Nina2017,Chen2017}．
%
%
対象がNNでなくともAEを生成できる\cite{Papernot2017}．
一方で，対象の勾配情報を利用できないことから，more challengingである．
}

{\bf 4. Using Discrete Cosine Transform (DCT) to perturb images:}
%
%
Naive formulation of the AE design problem enlarges the problem size.
Therefore, we propose a DCT-based perturbation generation method to
suppress increase of the number of dimensions.
The concurrent work~\cite{Guo2019} also proposed a DCT-based
perturbation pattern design method; however, this method optimizes single
objective function.
%

\subsection{Formulation}

\subsubsection{Design Variables}



In the proposed method, there are two methods to determine how to
perturb an input image: direct and DCT-based method.
\begin{itemize}
 \item {\bf Direct method:}
	   In the direct method, pixel intensity of input image $\Vec{I}$ is
	   perturbed directly based on a solution candidate $\Vec{x}$.
	   Thus, $\Vec{x}$ comprises variables $x_{u,v,c}^{(Dir)}$ as follows:
	   \begin{equation}
%
		\Vec{x} =\left\{x_{u,v,c}^{(Dir)}\right\}_{(u,v,c) \in \Vec{I}} 
	   \end{equation}
	   where $(u,v)$ denotes a $N_w \times N_w$ pixels block position in
	   $\Vec{I}$, and $c$ denotes color components.
	   The resolution of $\Vec{I}$ is $W_{\Vec{I}} \times H_{\Vec{I}}$
	   pixels and $\Vec{I}$ is decomposed into 
	   $\lceil \frac{W_{\Vec{I}}}{N_w} \rceil 
        \times 
        \lceil \frac{H_{\Vec{I}}}{N_w} \rceil $ blocks.


%
%

 \item {\bf DCT-based method:}
	   When generating adversarial examples for high resolution images,
	   the direct method requires many variables and the problem becomes
	   huge. 
	   Therefore, this study proposes an alternative method using two
	   dimensional Discrete Cosine Transform (2D-DCT), which is called
	   as a DCT-based method.
	   The DCT-based method involves two types of variables as follows:
	   \begin{eqnarray}
		\Vec{x} &=& \Vec{\chi} \cup \Vec{x}^{(DCT)}
		\\ 
		\Vec{\chi} &=&  \left\{ \chi_{u,v}^{(PS)}  \right\}_{(u,v) \in \Vec{I}} 
		\\ 
%
		\Vec{x}^{(DCT)} &=&  \left\{\Vec{x}^{(DCT)}_r \right\}_{1 \leq r \leq N_{AP}}
		\\ 
		\Vec{x}^{(DCT)}_r &=& \left\{ x_{p,q,r}^{(DCT)} \right\}_{1 \leq p \leq N_{DCT}, 1 \leq q \leq N_{DCT}}
	   \end{eqnarray}
	   where $\Vec{x}_{p,q,r}^{(DCT)}$ represents
	   alteration
	   pattern of 2D-DCT coefficients of subband $(p,q)$.
	   To adaptively perturb input image $\Vec{I}$ according to image
	   block features, the DCT-based method prepares $N_{AP}$ alteration
	   patterns and suffix $r$ represents the pattern index.
	   $\chi_{u,v}^{(PS)}$ determines the generated 2D-DCT coefficient
	   alteration patterns to apply image block $(u,v)$ in input image
	   $\Vec{I}$, i.e., $\chi_{u,v}^{(PS)} = 0, 1, \ldots, N_{AP}$.
	   If $\chi_{u,v}^{(PS)} > 0$, the corresponding alteration pattern
	   is applied to block $(u,v)$, otherwise, the frequency coefficients
	   of the block do not change.
%

%
	   


\end{itemize}

\subsubsection{Objective Functions}

\del{
The AE design problem essentially involves more than one objective
function that have trade-off relationship, e.g., classification accuracy
and visibility of pertubation pattern.
Therefore, it is natural to formulate the problem as multi-objective
optimization and solve it by simultaneously optimize the objective
functions without gathering them into single function that require a
weight parameter.
}

In this paper, the following three scenarios are considered to
demonstrate the advantage of the proposed EMO-based approach.

\del{%
本研究では，AEの生成を多目的最適化問題としてモデル化することのメリットを
示すために，特に，様々なAEの作り方が可能となることを示すため，いくつかの
シナリオを検討する．
}

\begin{itemize}
 \item {\bf Accuracy versus perturbation amount scenario}

This is the fundamental scenario of multi-objective adversarial example
	   generation including the following two objective functions:
\begin{eqnarray}
{\rm minimize}&& f_1 = P( \mathcal{C}(\Vec{I} + \Vec{\rho} ) = \mathcal{C}(\Vec{I}))
\nonumber \\
{\rm minimize}&& f_2 =  
            || \Vec{\rho} ||_e
\end{eqnarray}
The first objective function $f_1$ indicates a probability that a target
classifier classifies a perturbed image $\Vec{I} + \Vec{\rho}$ to
the correct class $\mathcal{C}(\Vec{I})$ where
$\mathcal{C}(\cdot)$ denotes a classification result.
The second objective function indicates the amount of the perturbation
$\Vec{\rho}$ which can basically be calculated by $l_e$ norm of $\Vec{\rho}$.
This scenario clarifies the trade-off relationship between the
classification accuracy and the perturbation amount while generating various 
perturbation patterns.
%

\del{
摂動量と誤認識率とのトレードオフ関係上にあるパレート解を求めるシナリオで
ある．
摂動量と誤認識率との関係性を明確にしつつ，トレードオフ上にある多様なAEを
同時に生成できる．
}

\item {\bf $l_0$ versus $l_1$ norms scenario}

The gradient-based method generates AE by giving small perturbation to
	   all pixels of a target
image, and EC-based previous work~\cite{Su2017} generates AEs by perturbing
	   one or relatively small number of pixels.
On the other hand, the proposed method can comprehensively generate 
various AEs that have different number of perturbed pixels located between
AEs generated by gradient- and EC-based methods.
To this end, the number of perturbed pixels is employed as one of
objective functions.
The followings are example objective functions: 
\begin{eqnarray}
{\rm minimize}~ && f_1(\Vec{x}) =   || \Vec{\rho} ||_0
\nonumber \\
{\rm minimize}~ && f_2(\Vec{x}) =   || \Vec{\rho} ||_1
\nonumber\\
{\rm subject~to} && P( \mathcal{C}(\Vec{I} + \Vec{\rho} ) = \mathcal{C}(\Vec{I})) < T_{acc}
\end{eqnarray}
where $|| \vec{\rho} ||_0$ denotes the number of pixels whose values are
not zero in $\Vec{\rho}$, 
and $T_{acc}$ is a threshold.

\del{
FSGMは全画素に対して微小な摂動を与えることでAEを生成する． また，
ECベースのAE生成は，1画素またはごく少数の画素に対してのみ摂動を与えるこ
とでAEを生成する．
これに対して提案手法は，$f_1$を変更を加える画素数，$f_2$を各画素あたりの
階調値の変更幅とし，ともに最小化を行うことにより，
FGSMやECベースの手法の中間的なAEを網羅的に作成することが可能となる．
ここで
$| \rho_{\neq 0} |$は，$\rho$において画素が0以外，すなわち，摂動を加える
	   画素数を表す．
}

 \item {\bf Robust AE generation scenario}
%
Robust optimization is one of the optimizations taking advantage of the
	   characteristics of evolutionary computation~\cite{Shimoyama2007,Ono2009multi}.
Previous work was based on white-box
setting~\cite{Shin2017,Eykholt2018,Athalye2018} and
minimizes only averaged (or expected) classification
accuracy\cite{Athalye2018,Eykholt2018}; however,
this might cause AEs that could be correctly classified under a certain
condition because such rare cases cannot be represented the averaged
value.
Adding deviation to objective functions prevents such exceptional
failure of misclassification, resulting in generating more robust
AEs against image transformation.
\begin{eqnarray}
{\rm minimize}~ &&f_1(\Vec{x})  = 
%
				    \mathbb{E}\left( 
					   P( \mathcal{C} \left( \tau_i( \Vec{I} + \Vec{\rho} ) \right) 
  					      = \mathcal{C}(\Vec{I}))
			        \right)
\nonumber \\
{\rm minimize}~ &&f_2(\Vec{x}) = 
                   \sigma\left(
					   P( \mathcal{C} \left( \tau_i( \Vec{I} + \Vec{\rho} ) \right) 
  					      = \mathcal{C}(\Vec{I}))
				   \right)
\nonumber \\
{\rm minimize}~ &&f_3(\Vec{x}) =   || \Vec{\rho} ||_e
\nonumber \\
\end{eqnarray}
where $\mathbb{E}(\cdot)$ and $\sigma(\cdot)$ are expected value and standard
	   deviation of
classification accuracy, and
$\tau_i(\cdot)$ denotes image transformation.
%

\end{itemize}

Note that these three scenarios
have different purposes from each other
but share the need for multi-objective
optimization.

\subsection{Process Flow}

The proposed algorithm adopts any evolutionary multi-objective
optimization algorithms such as NSGA-II\cite{Deb2002} and
MOEA/D\cite{Zhang2007b}.
Here we explain the process flow of the proposed method taking MOEA/D as
an example.

MOEA/D converts the approximation problem of the true Pareto Front into
a set of single-objective optimization problems.
Here, an original multi-objective optimization problem is described as  follows:
\begin{eqnarray}
 {\rm minimize} && \Vec{f}(\Vec{x}) = \left( f_1(\Vec{x}), \ldots f_{N_f} (\Vec{x}) \right)
  \nonumber \\
 {\rm subject~to} && \Vec{x} \in \Vec{\mathcal{F}}
\end{eqnarray}
There are several models to convert the above problems into scalar
optimization problems;
%
for instance, in Tchevycheff approach, the above problem can be
decomposed into the following problem.
\begin{eqnarray}
 {\rm minimize} && g(\Vec{x} | \Vec{\lambda}^j, z^*) 
                   = \max_{1 \leq i \leq N_f}  
                     \left\{ \lambda_i^j | f_i(\Vec{x}) - z_i^*  \right\}
  \nonumber \\
 {\rm subject~to} && \Vec{x} \in \Vec{\mathcal{F}}
 \end{eqnarray}
where $\Vec{\lambda}^j = (\lambda_1^j, \ldots, \lambda_{N_f}^j )$ are weight
vectors ($\lambda_i^j \geq 0$
) and 
$\sum_{i=1}^{N_f} \lambda_i^j = 1$, and 
$\Vec{z}^*$ is a reference point calculated as follows:
\begin{equation}
z_i^* = \min\{ f_i(\Vec{x}) |  x \in \Vec{\mathcal{F}} \} 
\label{eq:reference_point}
\end{equation}
%
%
By preparing $N_D$ weight vectors and optimizing $N_D$ scalar objective
functions, MOEA/D finds various non-dominated solutions at one
optimization.

The detailed algorithm of the proposed method based on MOEA/D is as
follows:
%

\hspace*{-2ex}{\bf [Step 1] Initialization}

\hspace*{-2ex}{\bf [Step 1-1] Determine neighborhood relations for each
weight vector $\Vec{\lambda}^i$.}
By calculating the Euclidean distance between weight vectors, $N_n$
neighboring weight vectors $\{ \lambda^k \}$ ($k \in B(i) = \{ i_1,
\ldots, i_{N_n} \} $) are selected.

\hspace*{-2ex}{\bf [Step 1-2] Generate an initial population.}
The initial solution candidates $\Vec{x}_1, \ldots, \Vec{x}_{N_p - 1}$
are generated by sampling them at uniformly random from
$\Vec{\mathcal{F}}$.
The solution whose all variable values are set to 0, which corresponds
to involving no perturbation and would survive on the edge of the Pareto
front throughout the optimization, is also added to the initial
population.

\hspace*{-2ex}{\bf [Step 1-3] Determine the reference point.}
The reference point is calculated by eq.(\ref{eq:reference_point}).

\hspace*{-2ex}{\bf [Step 2] Selection}
$N_f$ best individuals are selected for $N_f$ objective functions
respectively, and then, by applying tournament selection, the indexes of
the subproblems $\mathcal{I}$ are selected ($\left| \mathcal{I} \right|
= \frac{N}{5} - N_f $).

\hspace*{-2ex}{\bf [Step 3] Population update}
The following steps 3-1 through 3-6 are conducted for each $i \in \mathcal{I}$.

\hspace*{-2ex}{\bf [Step 3-1] Selection of mating and update range.}
With the probability $\delta$, the update range $\mathcal{P}$ was limited to
$B{i}$, otherwise $\mathcal{P} = {1, \ldots, N_d}$.

\hspace*{-2ex}{\bf [Step 3-2] Crossover}
Randomly selects two indices $r_2$ and $r_3$ from $\mathcal{P}$ and set
$r_1 = i$, and generates a solution $\bar{\Vec{y}}$ whose element
$\bar{y}_k$ is calculated by the following equation:
\begin{equation}
\bar{y}_k = \left\{
\begin{array}{ll}
 x_k^{r_1} + F \left( x_k^{r_2} - x_k^{r_3} \right) & {\rm with~probability~} CR \\
 x_k^{r_1}                                          & {\rm with~probability~} 1 - CR \\
\end{array}
\right.
\end{equation}
The above equation is an operator proposed in
DE, and $CR$ and $F$ are control parameters.

\hspace*{-2ex}{\bf [Step 3-3] Mutation}
With the probability $p_m$, a polynomial mutation
operator~\cite{Deb1996} is applied to $\bar{\Vec{y}}$ to form a new candidate $\Vec{y}$
, i.e.,
the mutated value $y_k$ is calculated as follows: 
\begin{equation}
y_k = \bar{y}_k + \bar{\delta} \Delta_{max}
\end{equation}
where $\Delta_{max}$ represents the maximun permissible perturbance in
the parent value $\bar{y}_k$ and $\bar{\delta}$ is calculated as follows:
\begin{equation}
\bar{\delta} = \left\{
 \begin{array}{ll}
  (2 u)^\frac{1}{n + 1} - 1       & {\rm if}~ u < 0.5 \\
  1 - [2 (1 - u)]^\frac{1}{n + 1}  & {\rm otherwise}
 \end{array}			
\right.
\end{equation}
where $u$ is a random number in $[0, 1]$.

\hspace*{-2ex}{\bf [Step 3-4] Evaluation}
Evaluate $\Vec{y}$ by generating perturbation pattern $\Vec{\rho}$.

In the direct method, intensity $\rho_{a,b}$ at position $(a, b)$ of
perturbation pattern $\Vec{\rho}$ is directly determined by variables, i.e.,
\begin{equation}
\rho_{a,b} = x_{u,v}^{(DCT)}   
\end{equation}
where 
%
$1 \leq a \leq \Vec{I}_W$, $1 \leq b \leq \Vec{I}_H$, 
$u = \lfloor a / c \rfloor$, and 
$v = \lfloor b / c \rfloor$.

In the DCT-based method, DCT is applied to input image $\Vec{I}$ and
coefficients of basis functions $\bar{X}_{p,q}$ are obtained.
Then, values of $x_{p,q,r}^{(DCT)}$ in $\Vec{x}$ are added to the
coefficients $\bar{X}_{p,q}$ as follows:
\begin{equation}
 X_{p,q} = \bar{X}_{p,q} + x_{p,q,r}^{(DCT)}
\end{equation}
where $r = \xi_{u,v}^{(PS)}$ and $(u,v) \in \Vec{I}$.
Finally inverted DCT is applied to $X_{p,q}$ to form a perturbed image
$\Vec{I} + \Vec{\rho}$.

After generating the perturbed image $\Vec{I} + \Vec{\rho}$, a target
classifier is applied to it and obtains its recognition result
$\mathcal{C}(\Vec{I} + \Vec{\rho})$ with a confidence score, which is
referred to calculate
objective functions or constraints.
Other objective functions and constraints are calculated based on
$\Vec{\rho}$ or $\Vec{I} + \Vec{\rho}$.

\hspace*{-2ex}{\bf [Step 3-5] Update of reference point}
If $z_j > f_j(\Vec{y})$ for each $j = 1, \ldots, N_f$, 
then replace the value of $z_j$ with $f_j(\Vec{y})$. 

\hspace*{-2ex}{\bf [Step 3-6] Update of solutions}
Perform the following procedure to update population.
{\renewcommand{\labelenumi}{(\arabic{enumi})}
\begin{enumerate}
 \item Set $c = 0$.

 \item If $c = n_r$ or $\mathcal{P}$ is empty, then go to
	   (4). Otherwise, pick an index $k$ from $\mathcal{P}$ at random.


 \item If any of the following conditions are satisfied, then replace
	   $\Vec{x}^k $ with $\Vec{y}$ and set $c = c + 1$.
%
%
%
	   \begin{equation}
		\Vec{y} \not\in \Vec{\mathcal{F}} \land 
		 \Vec{x}^k \not\in \Vec{\mathcal{F}} \land
		 vio(\Vec{y}) < vio(\Vec{x}^k)		
	   \end{equation}
	   \begin{equation}
		\Vec{y} \in \Vec{\mathcal{F}} \land 
		 \Vec{x}^k \not\in \Vec{\mathcal{F}}
	   \end{equation}
	   \begin{equation}
		\Vec{y} \in \Vec{\mathcal{F}} \land
		 \Vec{x}^k \in \Vec{\mathcal{F}} \land
		 g(\Vec{y} | \lambda^k, \Vec{z})
		 \leq g(\Vec{x}^k | \lambda^k, \Vec{z})
	   \end{equation}
	   where $vio(\cdot)$ denotes the amount of constraint
	   violations.

 \item Remove $k$ from $\mathcal{P}$ and go back to (2)
\end{enumerate}
}

\hspace*{-2ex}{\bf [Step 4] Stop condition}
After iterated
$N_g$ generations, the algorithm
stops the optimization.
Otherwise, go back to Step 2.

%

\section{Evaluation}

\subsection{Experimental Setup}

Four experiments were conducted to demonstrate the effectiveness of the
formulation of AE generation problem as multi-objective optimization.
Experiment 1 shows whether the proposed method generates various AEs under
$l_0$ versus $l_1$ norms scenario,
i.e., the first objective function is the number of perturbed pixels and
the second one is $||\Vec{\rho}||_1$, both of which should be minimized.
Experiment 2 demonstrates whether the proposed multi-objective black-box
optimization approach can generate adversarial examples robust against
image rotation. 
Experiment 3 compares the proposed two methods, the direct method and the
DCT based method on a higher resolution image.
Experiment 4 demonstrates some examples of adversarial attacks on
ImageNet-1000 data.

In all the experiments, MOEA/D was used.
To convert the multi-objective optimization problem into a set of scalar
optimization problems, Tchebysheff approach is adopted.
The neighborhood size $N_n$ was set to 10, $\delta = 0.8$ and $n_r = 1$,

In experiments 1 and 2, we prepare canonical CNN models that involve 
\begin{itemize}
 \item two sets of convolution layers with ReLU activation function,
	   pooling and dropout layers,
 \item a fully connected layer with ReLU activation function followed by
	   a dropout layer, and 
 \item output layer consisting of a fully connected layer with softmax
	   activation function.
\end{itemize}
The above network was trained with Adam~\cite{Kingma2014} using 45,000
labeled images in CIFAR-10.  The batch size and the number of epoch were
set to 128 and 10, respectively.
In experiments 3 and 4, VGG16~\cite{Simonyan2014}, which is a widely-used
classifier based on CNN, was adopted.
We used the pretrained VGG16 model implemented on Keras framework.

\del{
\begin{verbatim}
    def __init__(self, problem,
                 neighborhood_size = 10,
                 generator = RandomGenerator(),
                 variator = None,
                 delta = 0.8,
                 eta = 1,
                 update_utility = None,
                 weight_generator = random_weights,
                 scalarizing_function = chebyshev,
                 **kwargs):
\end{verbatim}
}

\subsection{Experiment 1: $l_0$ versus $l_1$ norms of the perturbation pattern}

\begin{figure}[t]
 \centering
   \includegraphics[width=25mm]{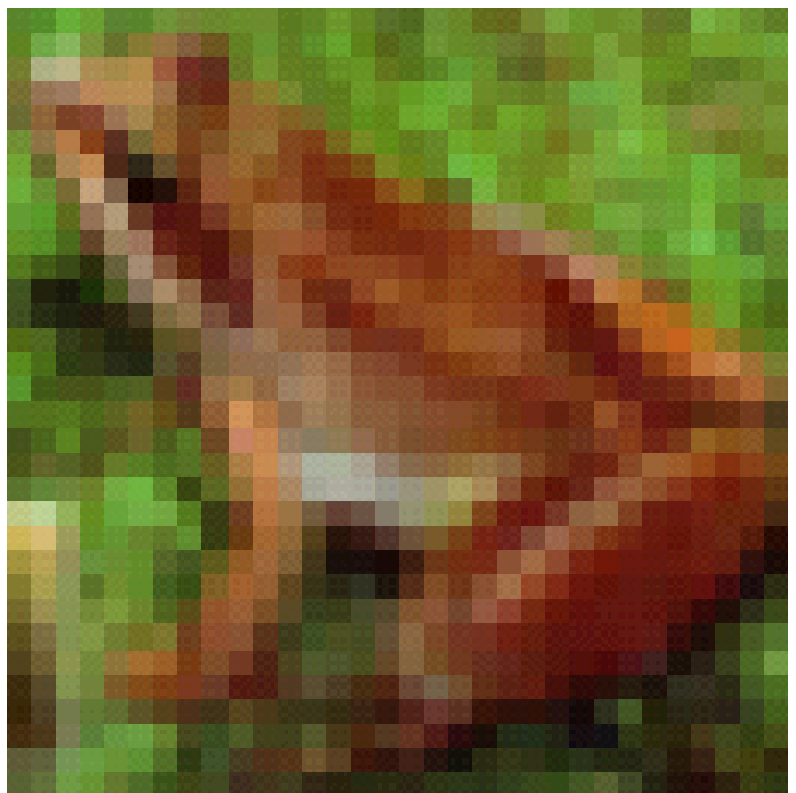}
 \caption{Input image $I_{1}$ used in experiments 1 and 2.}
 \label{fig:input_frog}
~\\
~\\
 \includegraphics[width=70mm]{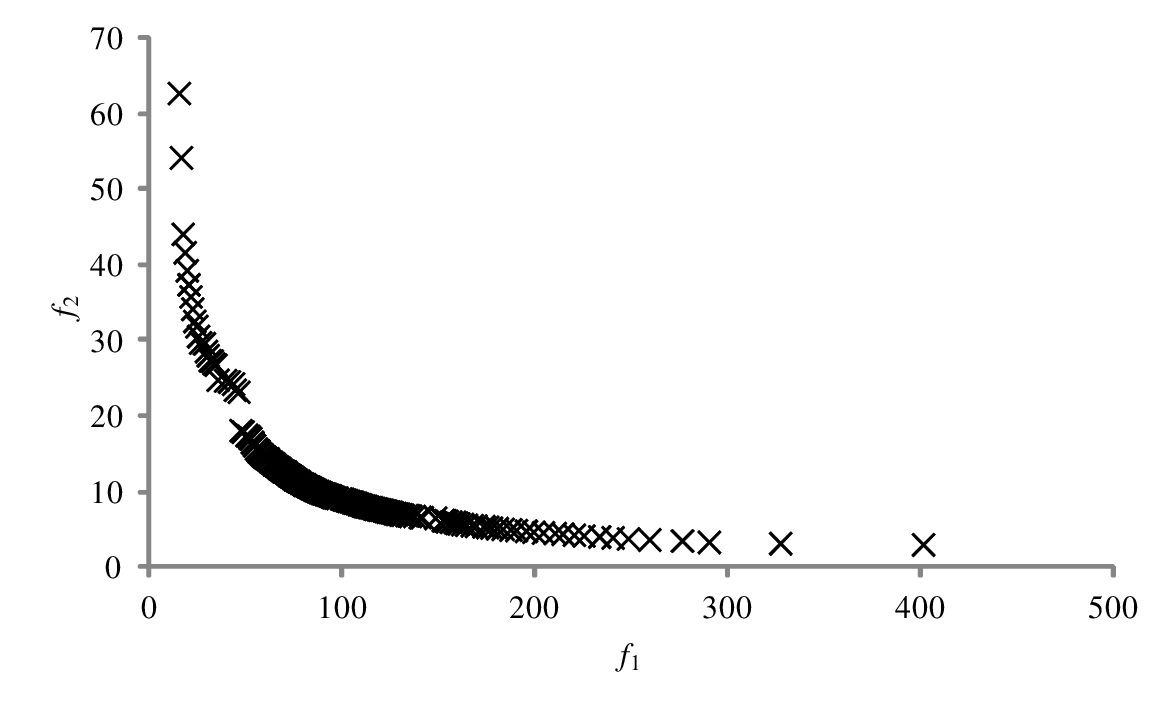}
 \caption{Results of experiment 1: obtained non-dominated soutions in $l_0$ versus $l_1$ norms scenario.}
 \label{fig:rst_pareto_num_vs_deg}
%
%
~\\
~\\
 \begin{tabular}[t]{ccc}
  \rotatebox{0}{\includegraphics[width=25mm]{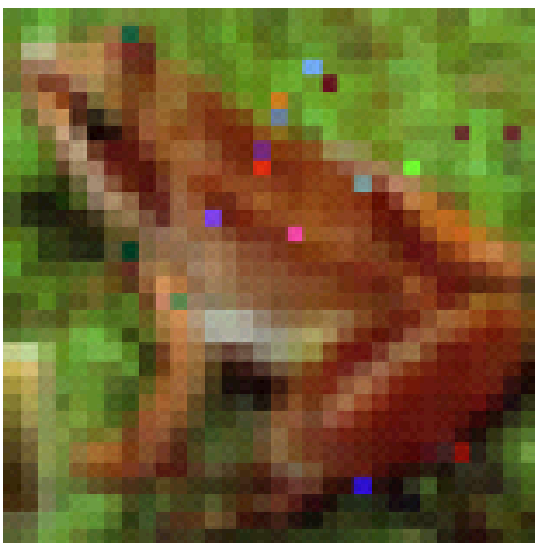}} &
  \rotatebox{0}{\includegraphics[width=25mm]{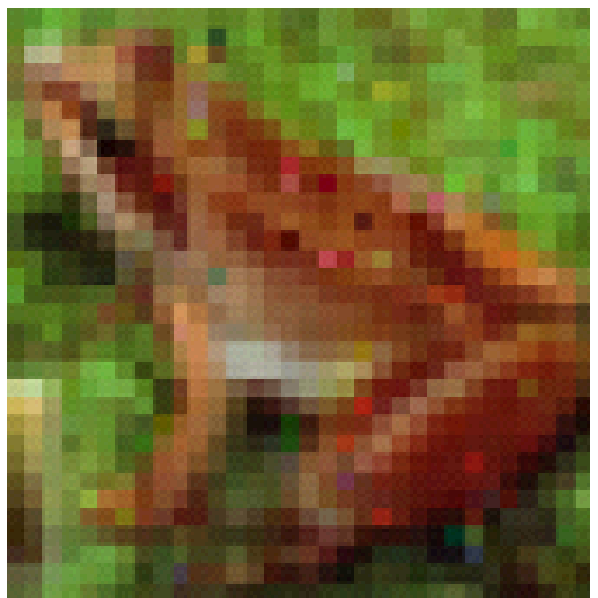}} &
  \rotatebox{0}{\includegraphics[width=25mm]{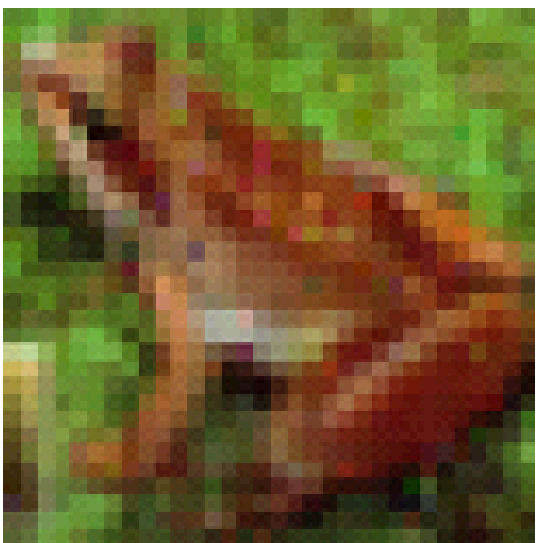}} \\
  \multicolumn{1}{p{25mm}}{\footnotesize i) $f_1 = 18$} & 
  \multicolumn{1}{p{25mm}}{\footnotesize ii) $f_1 = 100$} & 
  \multicolumn{1}{p{25mm}}{\footnotesize iii) $f_1 = 327$} \\
%
  \multicolumn{3}{c}{\footnotesize (a) Generated adversarial examlpes.}\\
%
  \rotatebox{0}{\includegraphics[width=25mm]{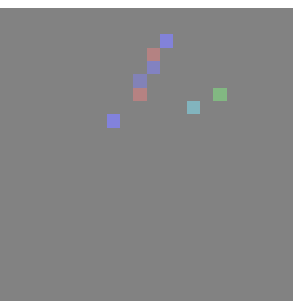}} &
  \rotatebox{0}{\includegraphics[width=25mm]{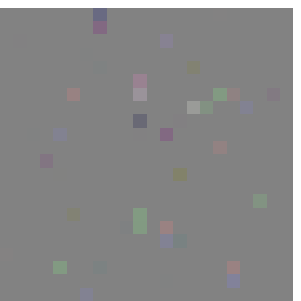}} &
  \rotatebox{0}{\includegraphics[width=25mm]{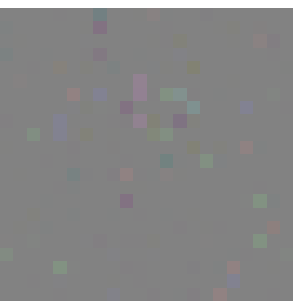}} \\
  \multicolumn{1}{p{25mm}}{\footnotesize i) $f_1 = 18$} & 
  \multicolumn{1}{p{25mm}}{\footnotesize ii) $f_1 = 100$} & 
  \multicolumn{1}{p{25mm}}{\footnotesize iii) $f_1 = 327$} \\
  \multicolumn{3}{c}{\footnotesize (b) Perturbed patterns.}\\
 \end{tabular}
 \caption{Results of experiment 1: generated adversarial examples in $l_0$ versus $l_1$ norm scenario.}
 \label{fig:rst_ae_image_num_vs_deg}
%
\end{figure}

In this first experiment, the proposed method was applied to design
adversarial examples for image $\Vec{I}_1$ shown in Fig.~\ref{fig:input_frog}
under 
$l_0$ versus $l_1$ norms scenario.
That is, the first objective function was the number of perturbed pixels
and the second objective function was the strength of changing pixel
intensity on the perturbed pattern $\rho$.
A constraint in which $P(\mathcal{C}(\Vec{I}_1 + \rho) =
\mathcal{C}(\Vec{I}_1))$ should be less than 0.2 was also considered.
The proposed method uses the direct method and set $N_w = 1$.
Because the input image size was $32 \times 32$ and they have 3 color
channels, the total number of design variables was 3,072. 
%
The population size and the generation limit were set to 500 and 1,000, respectively.

In this experiment, the initial population was generated by dividing
individuals into eight groups and imposing upper limits on the number of
pixels to be changed and pixel perturbation ranges.
Different upper limits were set for each group, 0.5\%, 5\%, 20\%, 35\%,
50\%, 65\%, 80\%, and 95\%, respectively, while pixel perturbation range
were also limited to $\pm 200$, $\pm 200$, $\pm 100$, $\pm 50$, $\pm
33$, $\pm 25$, $\pm 20$, and $\pm 16$, respectively.
The first two groups were also imposed to alter pixel values at least $\pm
150$ and $\pm 100$, respectively.

Fig.~\ref{fig:rst_pareto_num_vs_deg} shows the obtained non-dominated
solutions, which demonstrates that the proposed method could generate
various adversarial examples including ones in which 15 to over 400
pixels were changed and located between AEs generated by the previous
EC- and gradient-based methods.
%
%
%
Fig.~\ref{fig:rst_ae_image_num_vs_deg}(a) shows some examples of the
obtained by the proposed method.
All the three images shown in Fig.~\ref{fig:rst_ae_image_num_vs_deg}(a)
were classified to 'deer' with the confidence of 50.3\%, 45.9\%, and
41.8\%, respectively whereas originally, $I_1$ was classified to 'frog'
with the confidence of 99.28\%.
Fig.~\ref{fig:rst_ae_image_num_vs_deg}(b) shows the perturbation patterns
in which gray pixel indicates that were not modified, brighter pixels
represent that were changed to their intensity was increased, and darker
pixels represent that were changed in the opposite direction.
From the perturbation patterns shown in
Fig.~\ref{fig:rst_ae_image_num_vs_deg}(b), different perturbation
patterns could be seen, though similar distributions were observed
between ii) and iii).
%

\del{
\begin{verbatim}
20181116の結果: 
$f_1$は変更ピクセル数，
$f_2$はRMSE，
正答率は制約条件 50\%以下
1000個体5000世代

輝度値 vs 画素数（シナリオ1）
 - 20181102★
   - 初期集団工夫 ... いじる画素数をあらかじめ割合を決めて初期個体を生成
     - 1pixelだけいじるグループ
\end{verbatim}
}

\begin{figure}[t]
  \centering
  \hspace*{-10mm}
  \rotatebox{0}{\includegraphics[width=75mm]{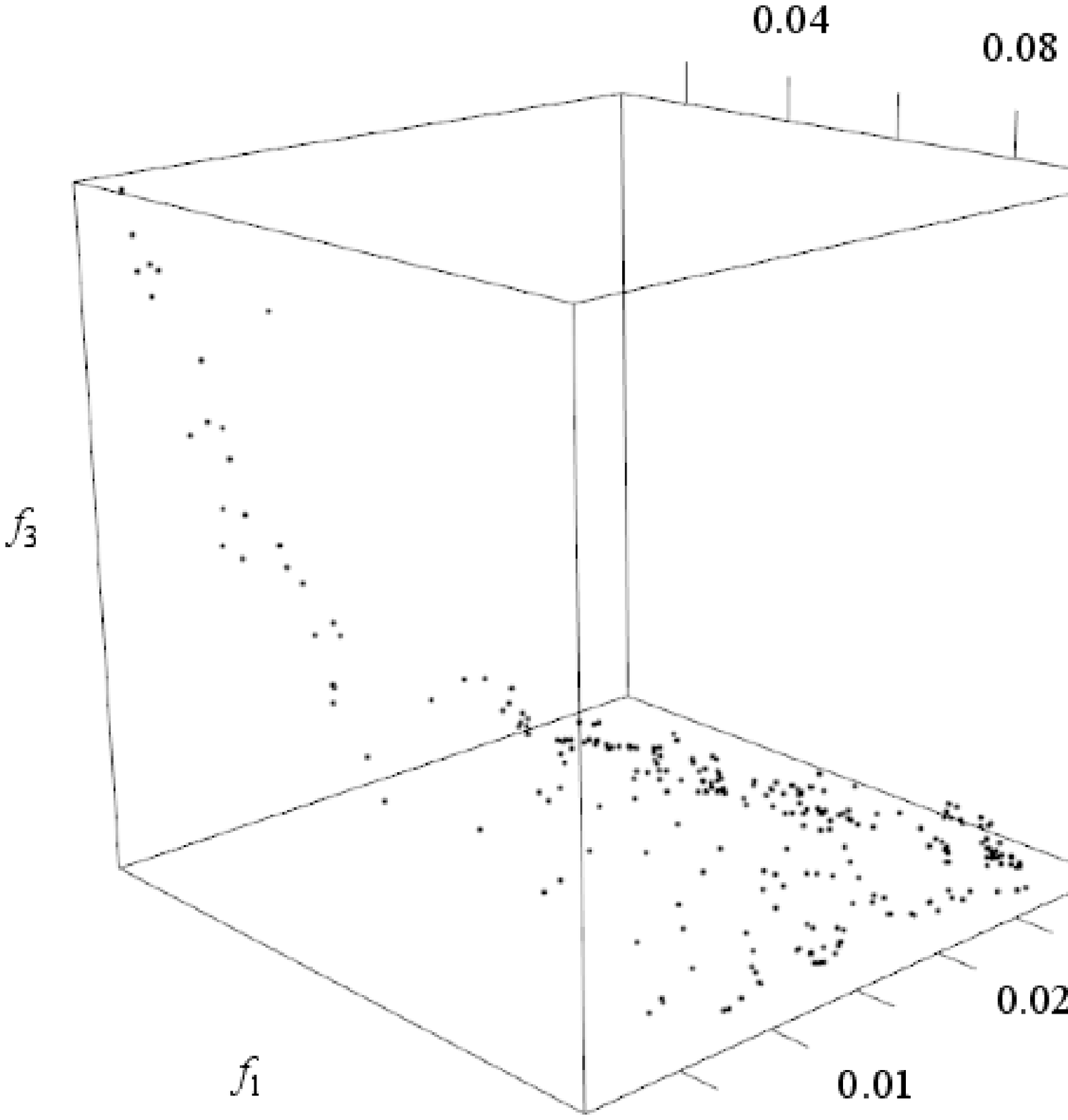}}
 \caption{Results of experiment 2: obtained non-dominated solutions for generating adversarial
 example robust against rotation.}
 \label{fig:rst_graph_pareto_robust}
~\\
~\\
 \begin{tabular}[t]{cc}
  \rotatebox{0}{\includegraphics[width=25mm]{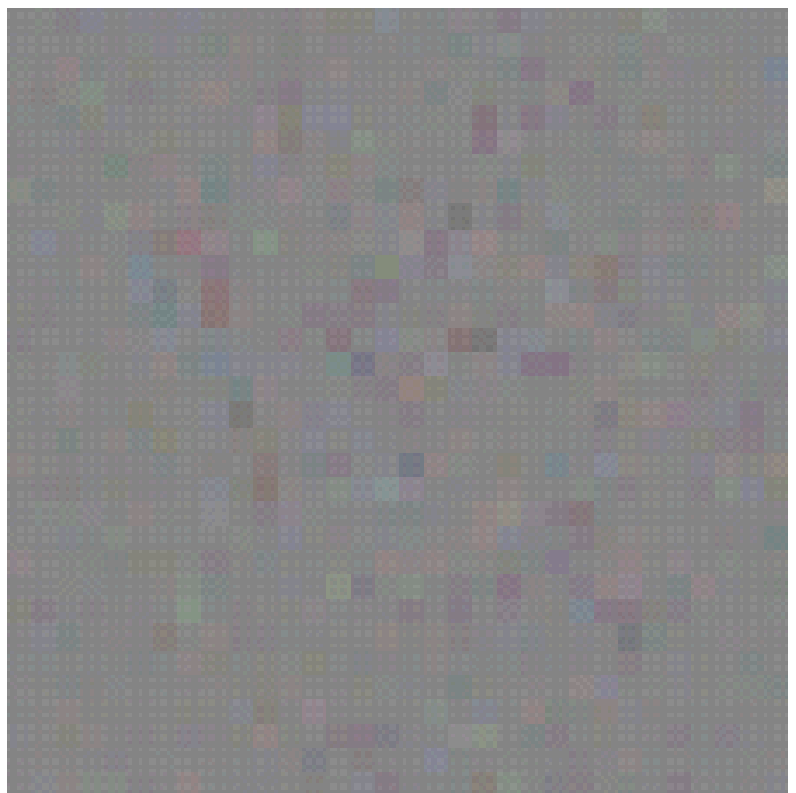}} &
  \rotatebox{0}{\includegraphics[width=25mm]{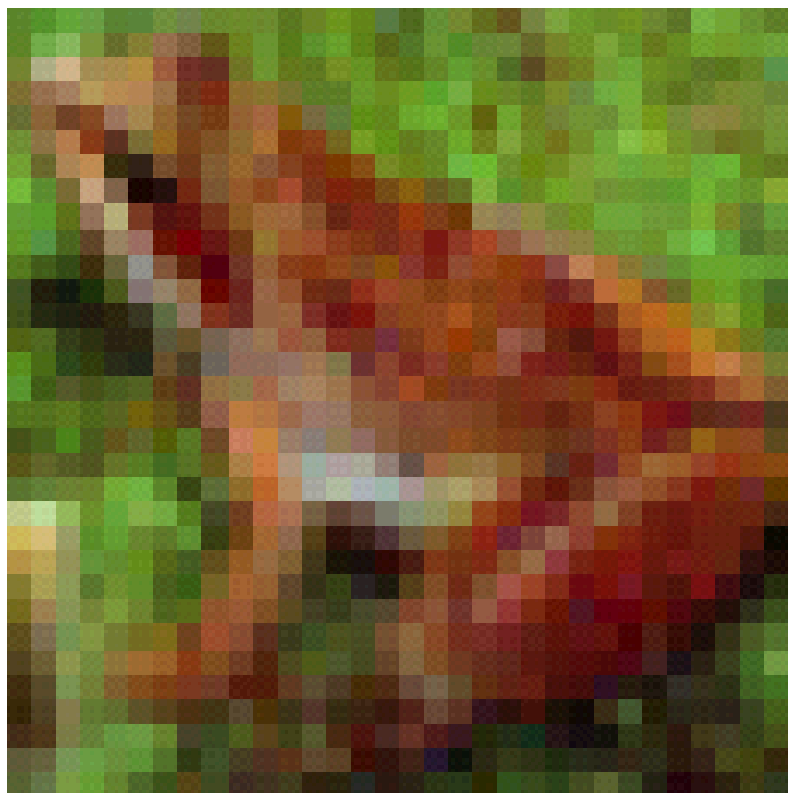}}\\
%
  \multicolumn{1}{p{25mm}}{\footnotesize (a) Perturbation pattern $\Vec{\rho}$ } &
  \multicolumn{1}{p{25mm}}{\footnotesize (b) Perturbed image $\Vec{I_1} + \Vec{\rho}$} \\
 \end{tabular}
 \caption{Results of experiment 2: generated adversarial example robust against rotation.}
 \label{fig:rst_robust_ae_image}
~\\
 
\includegraphics[width=60mm]{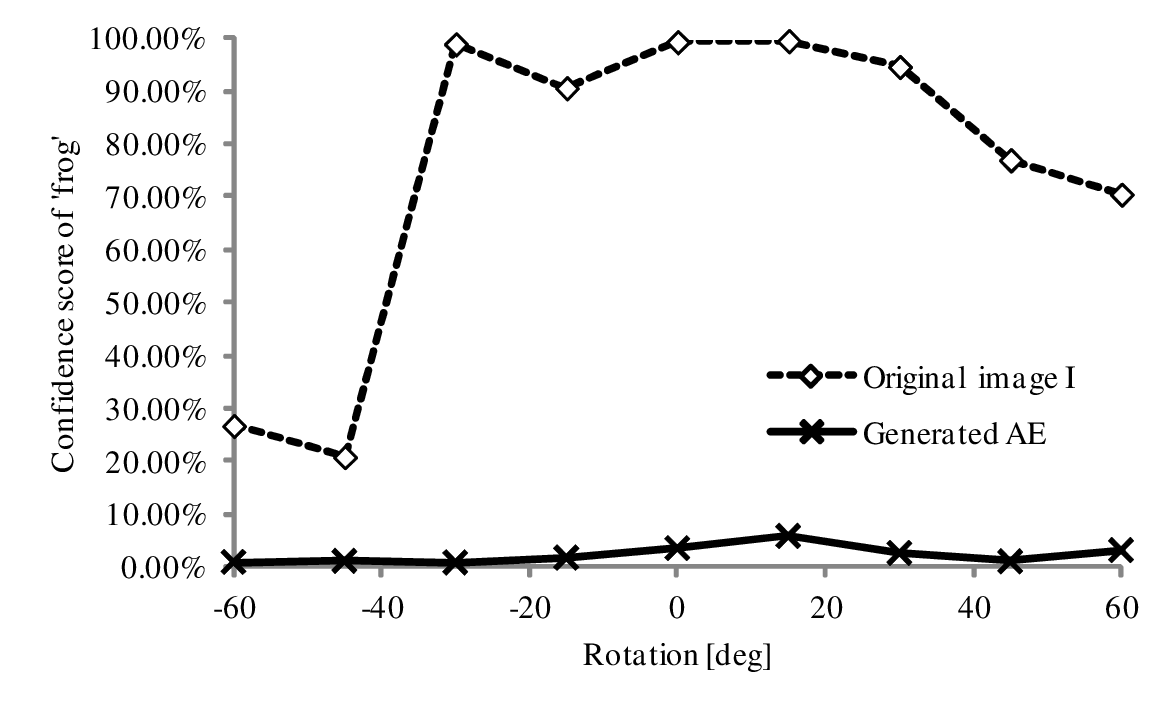}
\caption{Results of experiment 2: robustness of the generated example against rotation.}
\label{fig:rst_robust_ae_graph}

\end{figure}

\begin{table}[t]
 \caption{Results of experiment 2: classification results and confidence of the generated example
 robust against rotation.} 
 \label{tbl:rst_robust_ae}
 \centering
\begin{tabular}[t]{@{}l@{~}|@{~}l@{~}r@{~~}l@{~}r@{~}|@{~}l@{~}r@{~~}l@{~}r@{}}
\hline
  \multicolumn{1}{@{}c@{~}|@{~}}{Rotation}
& \multicolumn{8}{@{}c@{}}{Recognition results and confidence} 
\\   \cline{2-9}     
  \multicolumn{1}{@{}c@{~}|@{~}}{angle}
& \multicolumn{4}{@{}c@{~}|@{~}}{Clean image $I$}
& \multicolumn{4}{@{}c@{}}{Perturbed image $I + \rho$} \\
\hline
 -60 deg & Frog: & 26.8\% & (Cat: & 51.6\%) & Frog: & 1.0\% &(Cat:   &65.1\%) \\
 -45 deg & Frog: & 20.9\% & (Cat: & 69.0\%) & Frog: & 1.2\% &(Cat:   &69.3\%) \\
 -30 deg & Frog: & 98.9\% &       &         & Frog: & 0.9\% &(Truck: &96.3\%) \\
 -15 deg & Frog: & 90.6\% &       &         & Frog: & 1.8\% &(Bird:  &65.6\%) \\
   0 deg & Frog: & 99.3\% &       &         & Frog: & 3.6\% &(Deer:  &77.0\%) \\
  15 deg & Frog: & 99.5\% &       &         & Frog: & 5.9\% &(Truck: &44.0\%) \\
  30 deg & Frog: & 94.6\% &       &         & Frog: & 2.7\% &(Truck: &64.5\%) \\
  45 deg & Frog: & 77.0\% &       &         & Frog: & 1.1\% &(Cat:   &60.6\%) \\
  60 deg & Frog: & 70.5\% &       &         & Frog: & 3.2\% &(Cat:   &47.6\%) \\
\hline
\end{tabular}
\end{table}

\begin{figure*}[t]
 \centering
 \begin{tabular}{ccccc}
  \rotatebox{0}{\includegraphics[width=30mm]{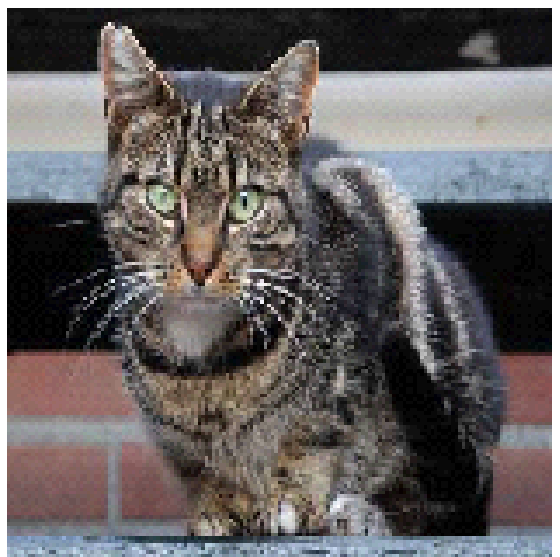}} &
  \rotatebox{0}{\includegraphics[width=30mm]{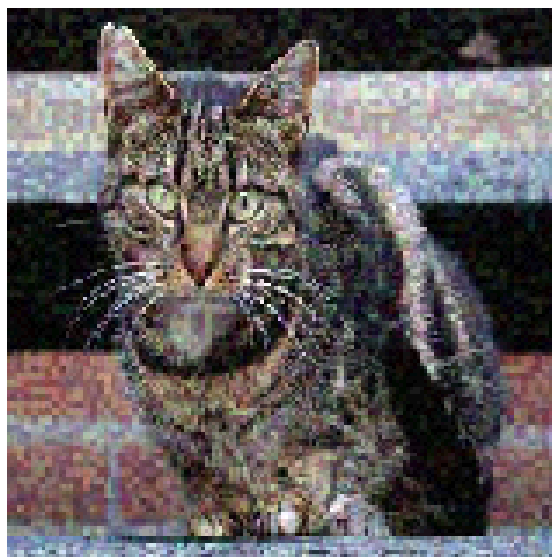}} &
  \rotatebox{0}{\includegraphics[width=30mm]{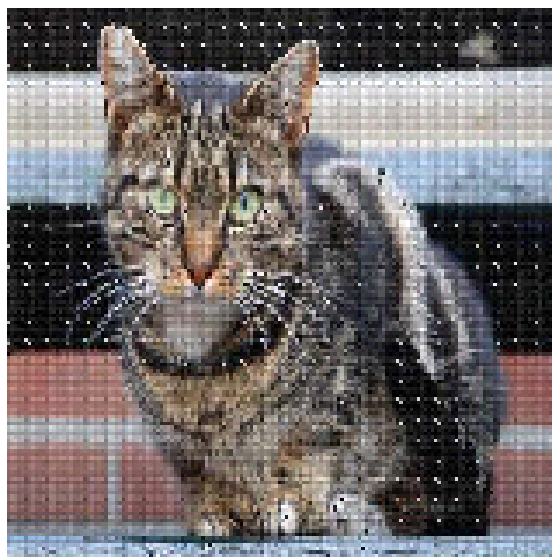}} &
  \rotatebox{0}{\includegraphics[width=30mm]{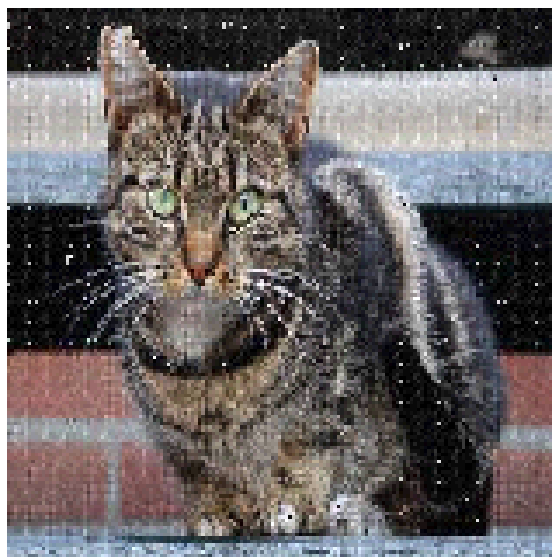}} &
  \rotatebox{0}{\includegraphics[width=30mm]{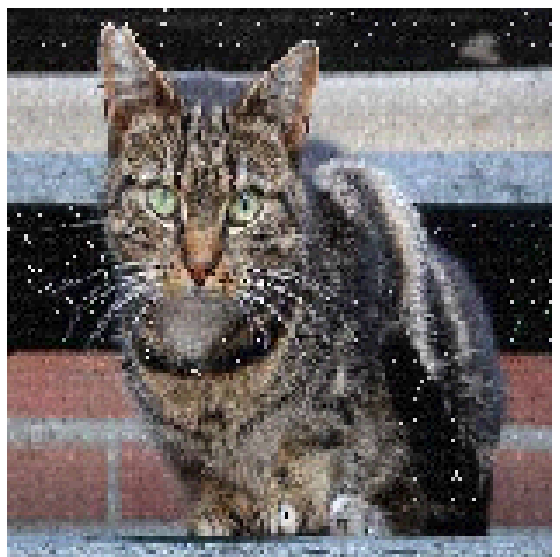}} \\
  \multicolumn{1}{p{25mm}}{\footnotesize (a) Original (clean) image $\Vec{I_2}$} &
  \multicolumn{1}{p{25mm}}{\footnotesize (b) Perturbed image $\Vec{I_2} + \Vec{\rho}$ by direct method} &
  \multicolumn{1}{p{25mm}}{\footnotesize (c) Perturbed image $\Vec{I_2} + \Vec{\rho}$ by DCT-based method ($N_{AP} = 1$)} &
  \multicolumn{1}{p{25mm}}{\footnotesize (d) Perturbed image $\Vec{I_2} + \Vec{\rho}$ by DCT-based method ($N_{AP} = 5$)} &
  \multicolumn{1}{p{25mm}}{\footnotesize (e) Perturbed image $\Vec{I_2} + \Vec{\rho}$ by DCT-based method ($N_{AP} = 10$)} \\
 \end{tabular}
 \caption{Results of experiment 3: generated adversarial examples by direct and DCT-based methods.}
 \label{fig:rst_highreso_ae_image}
\end{figure*}

\begin{table*}[t]
 \caption{Results of experiment 3: classification results and confidence of the generated example for VGG16.}
 \label{tbl:rst_robust_ae}
 \centering
\begin{tabular}[t]{@{}l@{~}|@{~}l@{~}r@{~}|@{~}l@{~}r@{~}|@{~}l@{~}r@{~}|@{~}l@{~}r@{~}|@{~}l@{~}r@{~~}l@{~}r@{~}}
\hline
%
  \multicolumn{1}{@{}c@{~}|@{~}}{}
& \multicolumn{2}{@{}c@{~}|@{~}}{}
& \multicolumn{8}{@{}c@{}}{Perturbed images $I + \rho$} \\
\cline{4-11}
  \multicolumn{1}{@{}c@{~}|@{~}}{Rank}
& \multicolumn{2}{@{}c@{~}|@{~}}{Clean image $I$}
& \multicolumn{2}{@{}c@{~}|@{~}}{Direct method}
& \multicolumn{6}{@{}c@{}}{DCT-based method}
\\ 
  \multicolumn{1}{@{}c@{~}|@{~}}{}
& \multicolumn{2}{@{}c@{~}|@{~}}{}
& \multicolumn{2}{@{}c@{~}|@{~}}{}
& \multicolumn{2}{@{}c@{~}}{$N_{AP} = 1$}
& \multicolumn{2}{@{}c@{~}}{$N_{AP} = 5$}
& \multicolumn{2}{@{}c@{}}{$N_{AP} = 10$}
\\ \hline
 1st & Tabby:        & 60.8\% & Envelope:      & 13.6\% & jigsaw\_puzzle: & 76.4\% & Purse:         &20.8\% &
       Coyote:       & 35.2\% \\
 2nd & Tiger\_cat:   & 30.4\% & Jigsaw\_puzzle:& 10.0\% & tabby:          & 4.6\% & Wood\_rabbit:  &13.2\% &
       Wallaby:      & 16.0\% \\
 3rd & Egyptian\_cat:&  7.4\% & Carton:        &  9.7\% & tiger\_cat:     & 2.8\% & Jigsaw\_puzzle: & 7.0\% &
       Wombat:       & 13.1\% \\
 4th & Doormat:      &  0.4\% & Wallet:        &  9.7\% & screen:         & 1.2\% & Window\_screen:& 6.9\% &
       Hare:         &  4.5\% \\
 5th & Radiator:     &  0.2\% & Door\_mat:     &  9.2\% & prayer\_rug:    & 1.1\% & Mitten:        & 6.9\% &
       German\_shepherd & 4.5\% \\
\hline
\end{tabular}
\end{table*}

\begin{table}[t]
 \centering
 \caption{Class labels regarded as correct ones in experiment 4.}
 \label{tbl:class_labels}
 \begin{tabular}[t]{@{}l@{~}|@{~}l@{~}|@{~}p{55mm}@{}}
  \hline
  Image & Original label & Labels regarded as correct \\
  \hline
  $I_3$ & Airliner   & Plane, Airship, Wing, Warplane, space\_shuttle \\
  $I_4$ & tiger\_cat & tabby, Egyptioan\_cat, Lynx, Persian\_cat, Siamese\_cat \\
  $I_5$ & electric\_guitar & acoustic\_guitar, Violin, Banjo, cello \\
  $I_6$ & Plastic\_bag & mailbag, sleeping\_bag \\
  $I_7$ & Promontory & Seashore, Lakeside, Cliff, cliff\_dwelling, Valley, Breakwater \\
  \hline
 \end{tabular}
\end{table}

\subsection{Experiment 2: generating robust AEs against image transformation}

In this experiment, taking the advantage of multi-objective
optimization, we attempt to design robust adversarial examples against
image transformation.
Simple image rotation was considered as image transformation in this
experiment because rotation has a greater influence than translation.
Here, for the purpose of enhancing the robustness against image
rotation, three objective functions were minimized: expected value and standard
deviation of recognition accuracy of transformed images ($f_1(\cdot)$
and $f_2(\cdot)$), and $l_1$ norm of perturbation pattern $\rho$
($f_3(\cdot)$).
Two constraints were also imposed: the recognition accuracy of the
target image was less than 10\% without rotation, and the 
expected
accuracy was less than 50\%.
The maximum rotation angle was set to $\pm 60$ degrees.
The population size and the generation limit were set to 500 and 2,000,
respectively.
Other experimental conditions were the same as experiment 1.

Fig.~\ref{fig:rst_graph_pareto_robust} shows the obtained non-dominated
solutions in the final generation.
We picked up one non-dominated solution from them and
Fig.~\ref{fig:rst_robust_ae_image} and
Fig.~\ref{fig:rst_robust_ae_graph} and show its image perturbation
pattern and its robustness against rotation, respectively.
The recognized class labels while changing rotation angle are shown in
Table \ref{tbl:rst_robust_ae}.
These results indicate that the generated AE successfully deceives the
classifier in both with or without rotation cases.

\del{

次に，多目的最適化の特性を活かして，頑健なAEの生成を試みた．
ここでは画像の回転操作に対する頑健性を高めることを目的として，○○
scenarioに基づいて3つの目的関数，すなわち，回転操作を適用された画像の認
識率の平均および標準偏差，および，摂動を加える輝度値幅のL1ノルムの3つを
最小化する問題とした．
また，制約条件として，正答率$P( \mathcal{C}(\Vec{x}) ) < 0.5$，回転画像
の平均正答率も50\%以下，すなわち，$\frac{1}{N_r} \sum_{i=1}^{N_r} P(
\mathcal{C}( \tau_i(\Vec{x}) ) ) < 0.5$とした．
対象画像の解像度が$32\times 32$pixelであり，RGBの3チャンネルの輝度値を
個別に変更することとした（変化幅は$\pm 32$）．
このため，次元数は3,072となり，
MOEA/Dのパラメータは，個体数を500，世代数を5,000として実験を行った．
対象はCIFAR-10のカエルの画像，および，●節で構築したCNNに基づく物体認識
モデルとした．
AEを生成する際に回転させた画像の角度は$\pm 60$度とした．

図●に得られたパレート解の分布を示す．
図の横軸は回転後の平均精度$f_1$，縦軸は$\rho$のL1ノルムを表す．
図より，輝度値の変更幅の平均が4未満の場合であっても，本来のクラスの
confideneceが10\%を下回るAEの生成に成功していることがわかる．
得られた非劣解のなかで，回転後の平均正答率が5.9\%，標準偏差が0.064，平均
輝度値変化が3.83のAEを図\ref{fig:rst_robust_ae_image}に示し，表
\ref{tbl:result_robust_ae}に回転角度毎の認識結果とその確率を示す．
図●より，●●であることがわかる．
また，表\ref{tb:rst_robust_ae}を見ると，摂動を加える前は，-45度以上の回
転を除いて，CNNが対象画像を正しくカエルと認識していたが，摂動を加えるこ
とで，-60度から60度の回転範囲において，カエルと認識する確率が30\%未満に
低下した．
以上のように，進化型多目的最適化問題を解くことによって，ブラックボックス
条件下，すなわち，モデルの内部情報を用いなくとも回転に対して頑健なAEを生
成できることがわかる．

\onote{【実験を仕掛け直せないか？ 制約条件を，カエルが1位にならないよう
に設定するとか．】}

\onote{【1目的最適化と比較してもいいかも？ 】}

}

\del{
\begin{verbatim}

×20181116

○20181123
$f_1$: 回転後の平均正答率，
$f_2$: 回転後の正答率の標準偏差，
$f_3$: 輝度値の平均，

制約: 回転後の正答率50\%以下，
変数: rgb輝度値（3072D），[-32～32]
500個体，5000世代

○20181129

$f_1$は変更ピクセル数，
$f_2$は平均輝度値，
$f_3$は回転後の正答率


○20181206

回転について:-45～45 度の範囲内で15 度ずつ回転して評価を行う．

第1 目的関数:回転後の平均正答率，
第2 目的関数:RMSE，
第3 目的関数:回転後の正答率の標準偏差

制約条件:正答率30\%以下、回転後の正答率40\%以下，
設計変数:DCT 係数?(64 変数),[-30～30] の範囲，
100 個体、8000 世代

	
robust AE @ cifar-10
 - 2～3日

 - 20181123の宝庫資料 直接法    -> こちら（の実験2）を掲載★
   - f1: ave acc
   - f2: stdev acc
   - f3:  （実験2のみ） 平均輝度値
   - const: acc < 50%
   - 500 pop, 5000 gen

 - 20181207の報告資料 DCT×
   - 一部明るいピクセル ... 範囲外？
\end{verbatim}
}

\subsection{Expleriment 3: effectiveness of the DCT-based method}

In order to verify the effectiveness of the DCT-based method in higher
resolution images, the direcet and DCT-based methods were compared on
generating AE for an image in ImageNet-1000 under accuracy versus
perturbation amount scenario.
The first objective function is the classification accuracy to the
original class.
In this experiment we consider more general class than the original label
assigned in ImageNet-1000, e.g., 
in the case generating AEs for image $\Vec{I}_2$ shown in
Fig.~\ref{fig:rst_highreso_ae_image}(a) which has a correct label `tabby',
labels of `Egyptian\_cat', `lynx', `Persian\_cat', `Siamese\_cat',
and `tiger\_cat' were also considered as correct labels.
The second objective function is Root Mean Square Error (RMSE) between
an original and perturbed images%
\footnote{%
The reason why we did not simply use $l_2$ norm of $\rho$ was to
evaluate the affection by DCT.
In the case using DCT-based method, the image quality slightly
deteriorated via DCT and inverse DCT even if the frequency coefficients
were not changed.
}.
A constraint, $P( \mathcal{C}(\Vec{I} + \Vec{\rho})) \leq 0.4$, was also
considered to enhance search exploitation.
In experiment 3 and subsequent experiments, we use the pretrained VGG16
as the target classifier.
%

In the case using the direct method, 
the total number of design variables was 5,625 because we changed the
input image resolution to $224 \times 224$, we set $N_w = 3$, and 
the perturbation was added to brightness component of $\Vec{I}$.
The DCT-based method requires less variables than the direct method,
i.e., 
$848$, $1,104$, and $1,424$ dimensions for $N_{AP} = 1$, $5$, and $10$,
respectively.

\del{
より高解像度の画像における有効性を評価するため，
Direct methodとDCT-based method
を用いてImageNet-1000におけるAEの生成を試みた．
対象とするモデルは学習済のVGG16とし\onote{【実装されているライブラリを明
記】}，画像は$224\times 224$にリサイズして与えることとした．
the first objective function $f_1$を正答率$P(\mathcal{C}(x))$（最小化），
the second objective function $f_2$をRMSE（最小化）とした．
また，制約条件としても正答率を考慮し，$P(\mathcal{C}(x)) \leq 0.4$とした。
%

Direct methodを用いる場合は，Cifar-10と比較して画像の解像度が高いため，
$3\times 3$画素を1つのblockとし，同じ量の輝度値変化を付与するものとした．
このため，設計変数の総数は$5,625$次元となった．

'Egyptian\_cat','lynx','Persian\_cat','Siamese\_cat','ta bby','tiger\_cat

DCTは$8\times 8$ pixelsのblock単位で適用するものとし，$N_{DCT}$パターン
のperturbationを用意することとした． 
また，対象画像のblockごとにperturbation patternを選択することとした．
このため，設計変数は$64 \times N_{DCT} + \left( 224 / 8 \right) ^ 2$とな
る．
$N_{DCT}$は1，5，10と変えてテストを行うこととする．
それぞれの場合の設計変数の総数は，●●，1,104，1,424次元となる．

Direct methodは50個体2,500世代，
DCT-based methodは個体数を100，世代数の上限
を4,000とした．
%

また，本実験で用いる学習済のVGG16は，●●の構造を持つCNNである．
ImageNet-1000を対象として，●●の学習を行ったモデルである．
}

Fig.~\ref{fig:rst_highreso_ae_image} shows the representative AEs
generated by the two methods,
and Table~\ref{tbl:rst_robust_ae} shows the recognition results and
confidence scores of the generated AEs.
Both methods could generate AEs that make the classifier misclassify.
In addition, the DCT-based method generated AEs including less conspicuous patterns
when $N_{AP} = 10$.

\del{

図●に示す猫の画像を対象とし，
Direct methodとDCT-based methodのパレート解の分布を図●に示す．
また，それぞれの結果におけるAEの例を図●に示す．
両方の手法において，誤認識率が高いAEを生成できることを確認した．
perturbationパターン数$N_{DCT}$を1，5，10と変更した場合，
$N_{DCT}$を増やすことにより，画像領域の特性に応じたperturbation pattern
の選択の自由度が増え，original imageに類似したAEを生成できていることがわ
かる．

一方，本実験では，図●に示すような，低いRMSEと低い正答率との双方を兼ね備えたAE
の生成に成功した一方で，ごく少ない非劣解に探索が集中してしまった．
多様なAEを生成したい場合は，$\epsilon$ dominance\cite{Laumanns2002}の導
入等を行う必要がある．

$N_{DCT} = 10$とし，その他の画像を対象としてAEを生成した結果を図●に示す．
●●であることがわかる．
%
\onote{【FGSMと比較できないか？】}

}

\begin{figure*}[t]
 \centering
 \begin{tabular}{@{}c@{~}c@{~}c@{~}c@{~}c@{}}
  \rotatebox{0}{\includegraphics[width=35mm]{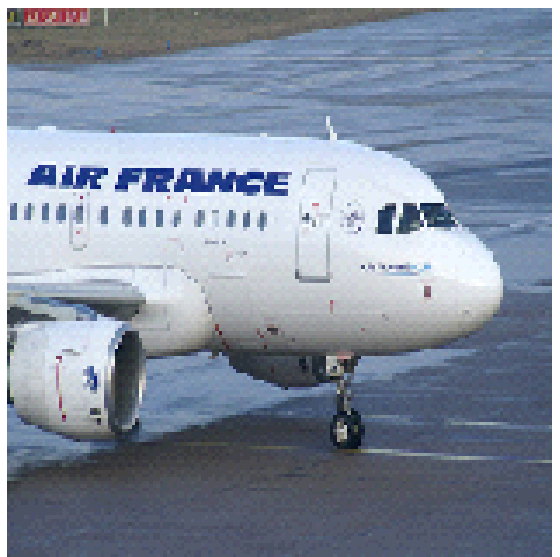}} &
  \rotatebox{0}{\includegraphics[width=35mm]{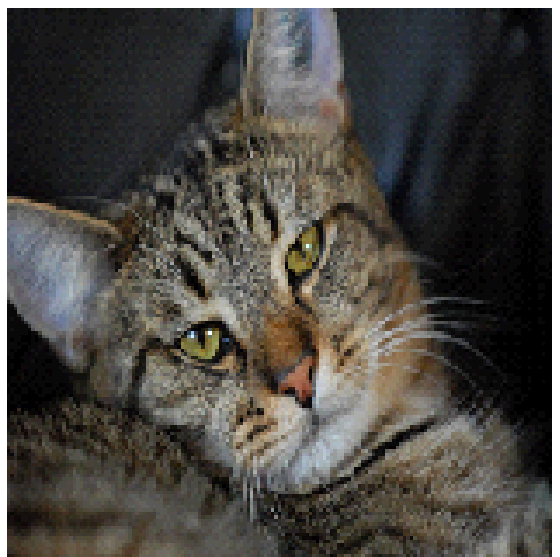}} &
  \rotatebox{0}{\includegraphics[width=35mm]{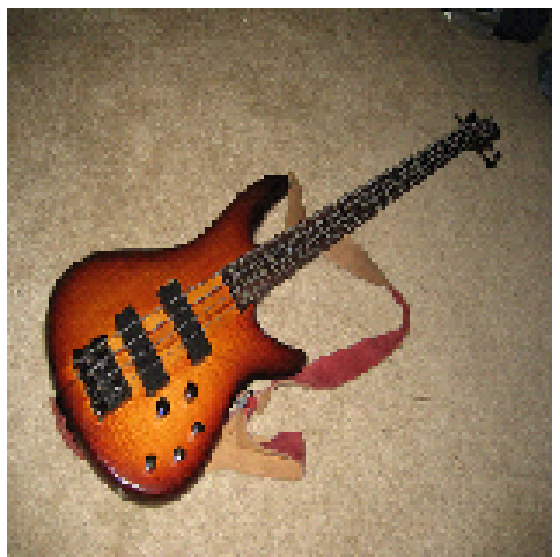}} &
  \rotatebox{0}{\includegraphics[width=35mm]{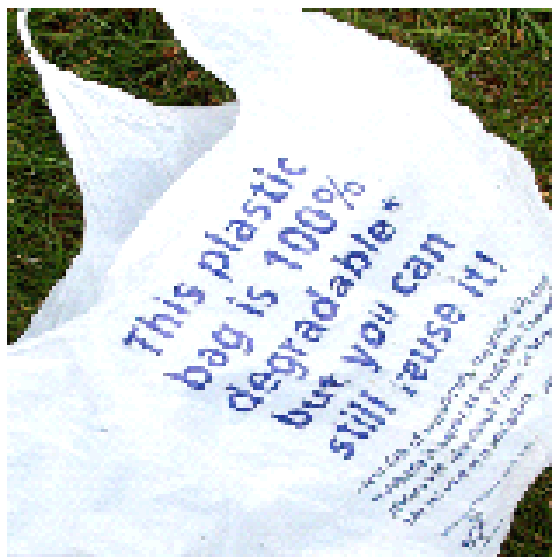}} &
  \rotatebox{0}{\includegraphics[width=35mm]{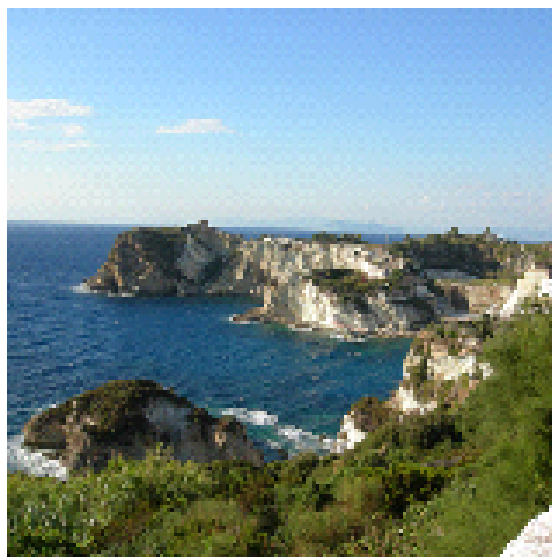}} \\
  $I_3$ &  $I_4$ &  $I_5$ &  $I_6$ &  $I_7$ \\
  \multicolumn{5}{c}{(a) Input clean images.}
  \\
%
  \hspace*{-5mm}\rotatebox{0}{\includegraphics[width=40mm]{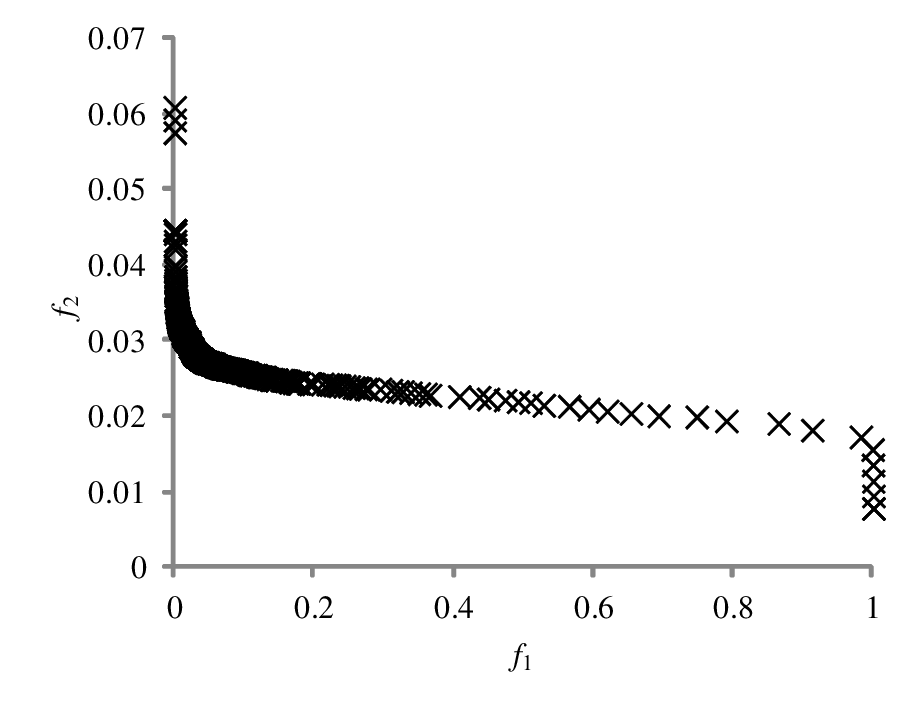}} &
  \hspace*{-5mm}\rotatebox{0}{\includegraphics[width=40mm]{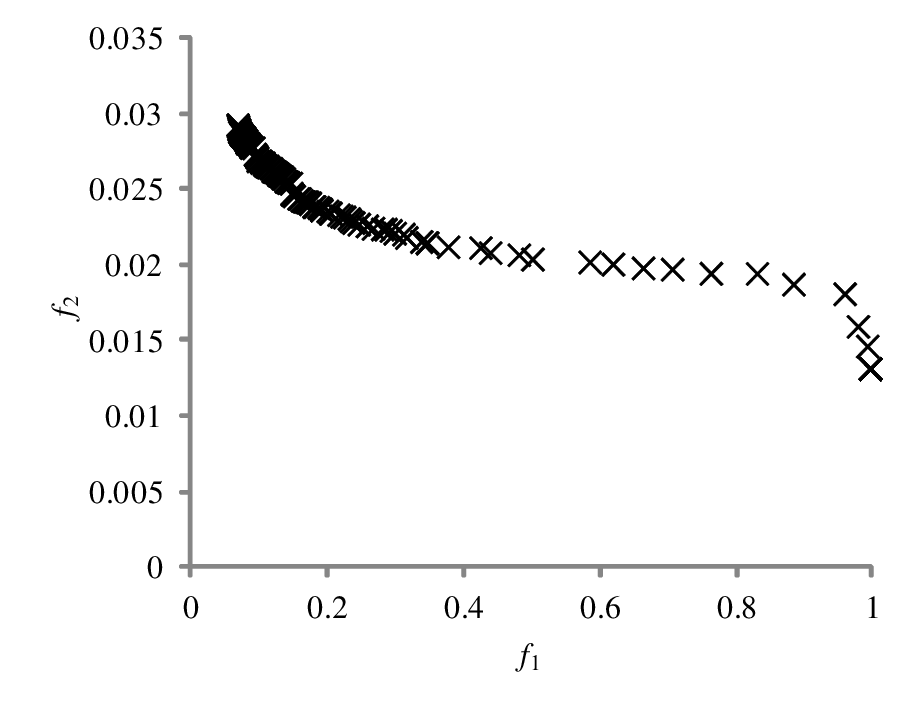}} &
  \hspace*{-5mm}\rotatebox{0}{\includegraphics[width=40mm]{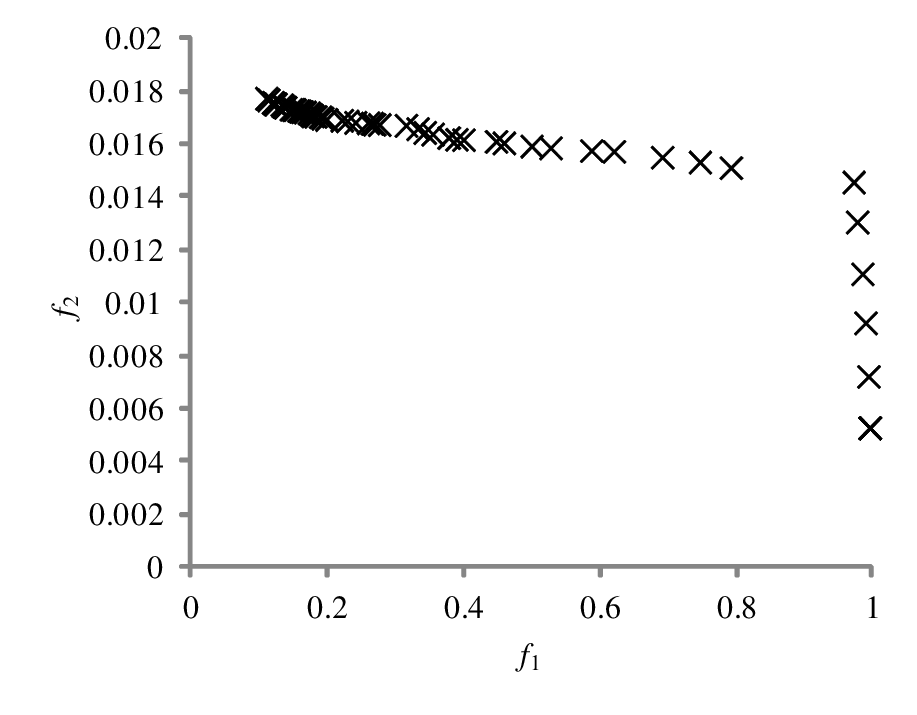}} &
  \hspace*{-5mm}\rotatebox{0}{\includegraphics[width=40mm]{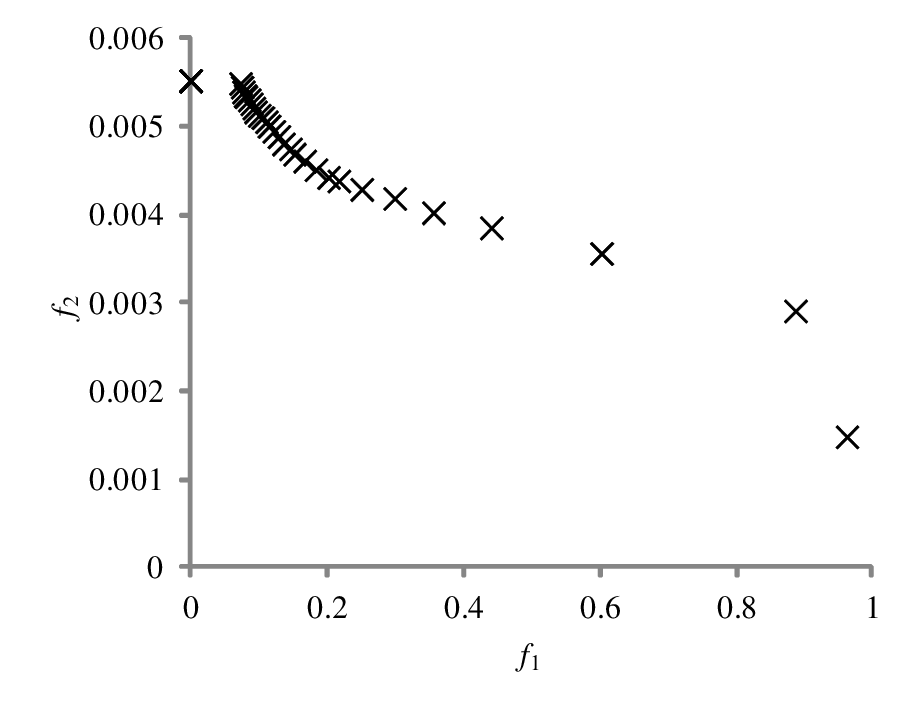}} &
  \hspace*{-5mm}\rotatebox{0}{\includegraphics[width=40mm]{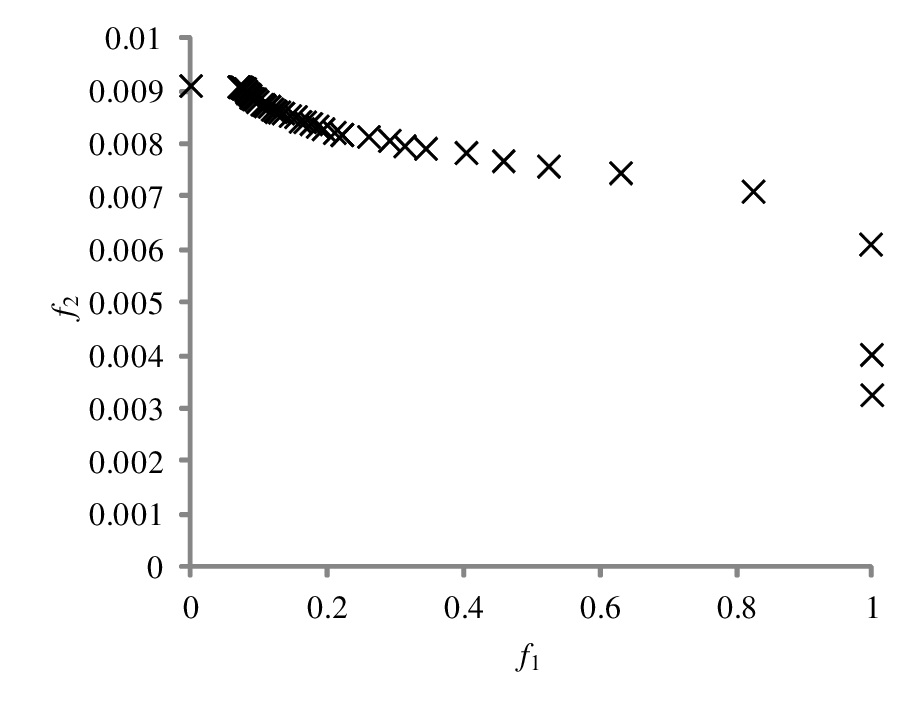}} \\
  $I_3$ &  $I_4$ &  $I_5$ &  $I_6$ &  $I_7$ \\
  \multicolumn{5}{c}{(b) Obtained non-dominated solutions.}
  \\
%
  \rotatebox{0}{\includegraphics[width=35mm]{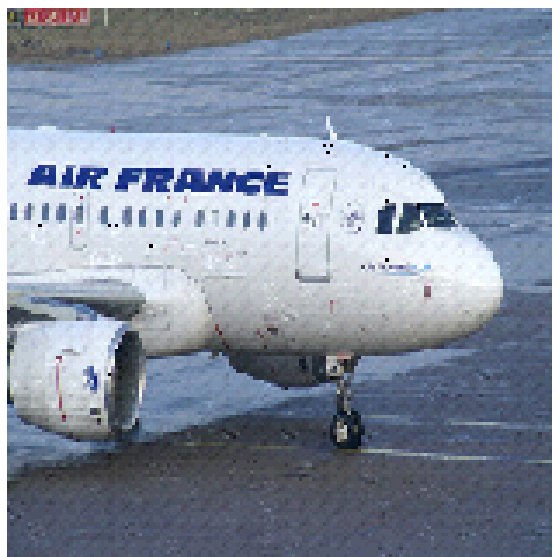}} &
  \rotatebox{0}{\includegraphics[width=35mm]{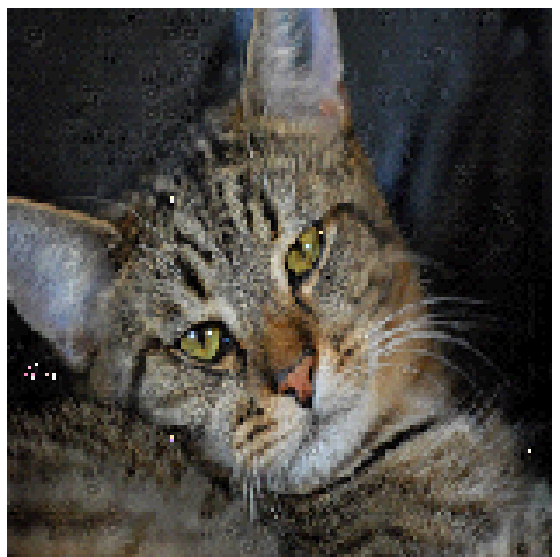}} &
  \rotatebox{0}{\includegraphics[width=35mm]{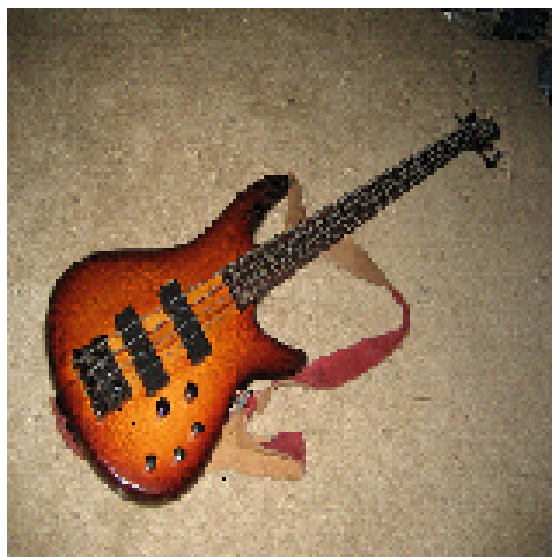}} &
  \rotatebox{0}{\includegraphics[width=35mm]{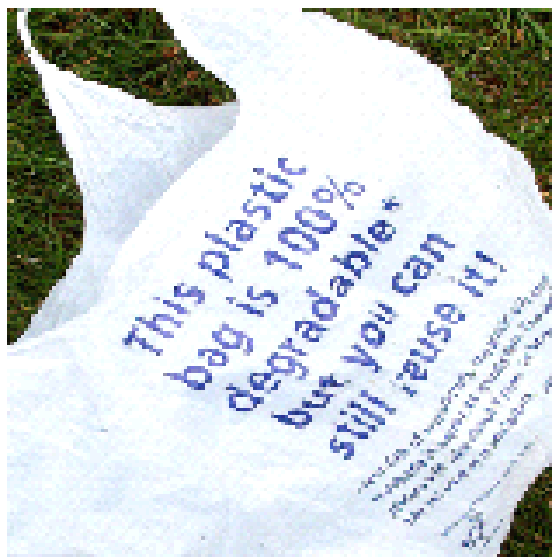}} &
  \rotatebox{0}{\includegraphics[width=35mm]{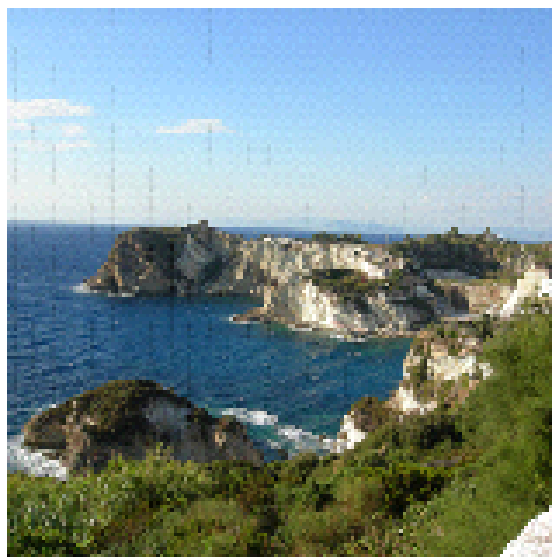}} \\
  $I_3$ &  $I_4$ &  $I_5$ &  $I_6$ &  $I_7$ \\
  \multicolumn{5}{c}{(c) Designed perturbation pattern examples.}
  \\
%
  \rotatebox{0}{\includegraphics[width=35mm]{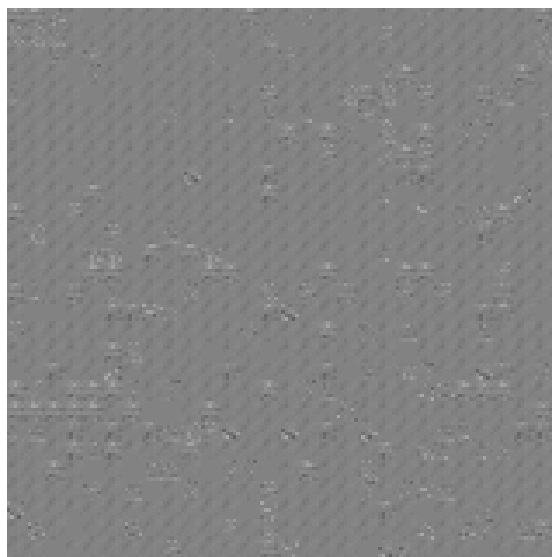}} &
  \rotatebox{0}{\includegraphics[width=35mm]{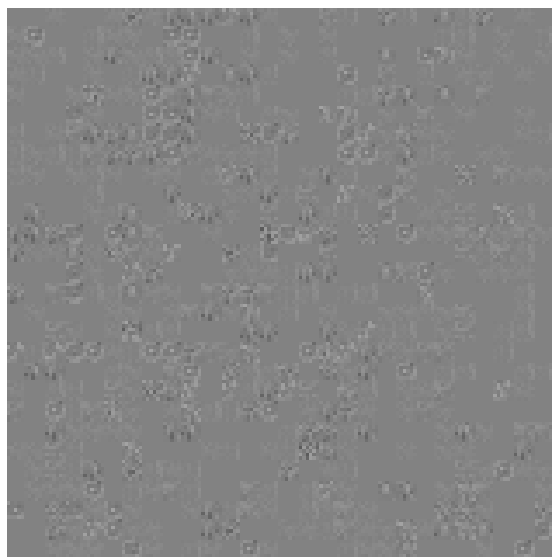}} &
  \rotatebox{0}{\includegraphics[width=35mm]{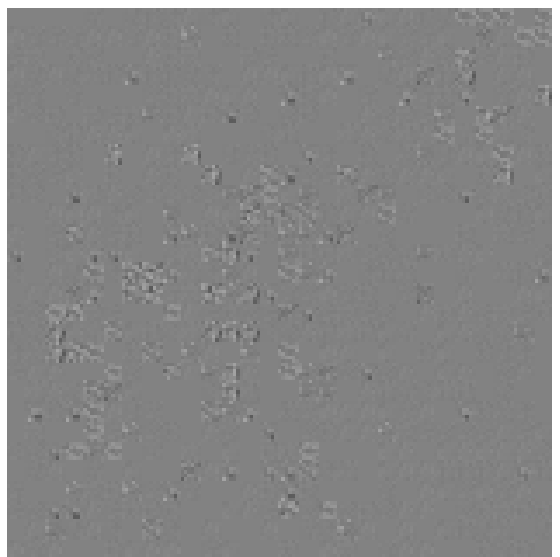}} &
  \rotatebox{0}{\includegraphics[width=35mm]{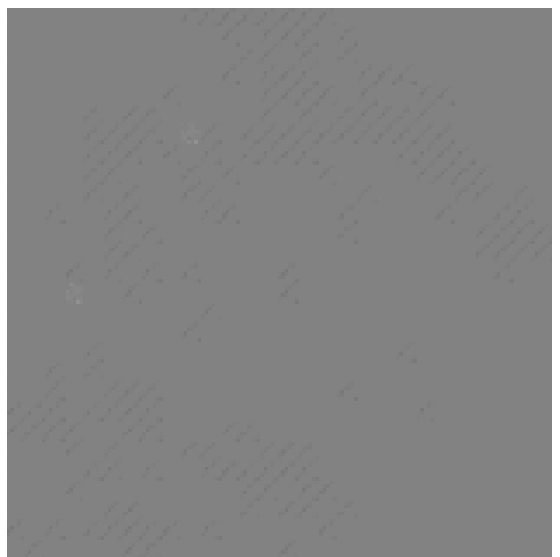}} &
  \rotatebox{0}{\includegraphics[width=35mm]{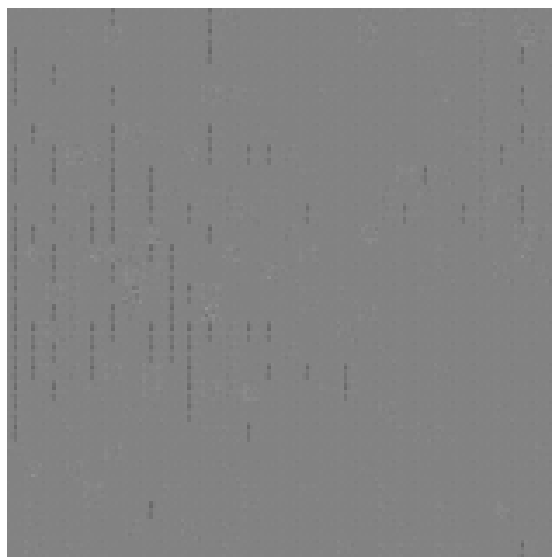}} \\
  $I_3$ &  $I_4$ &  $I_5$ &  $I_6$ &  $I_7$ \\
  \multicolumn{5}{c}{(d) Generated adversarial example images.}
  \\
%
 \end{tabular}

 \caption{Results of experiment 4: input images, obtained non-dominated solutions, designed
 perturb patterns, and generated adversarial example images.}
 \label{fig:rst_ae_image_dct}
\end{figure*}

\begin{table}[t]
 \centering
 \caption{Results of experiment 4: classification results and their confidence scores of original
 clean and perturbed images.}
 \label{tbl:rst_ae_image_dct}
 \begin{minipage}[b]{0.47\textwidth}
 \centering
  (a) $I_3$ \\
\begin{tabular}[t]{@{}l@{~}|@{~}l@{~}r@{~}|@{~}l@{~}r@{}}
\hline
  \multicolumn{1}{@{}c@{~}|@{~}}{Rank}
& \multicolumn{4}{@{}c@{}}{Recognition results and confidence} 
\\   \cline{2-5}     
  \multicolumn{1}{@{}c@{~}|@{~}}{}
& \multicolumn{2}{@{}c@{~}|@{~}}{$\mathcal{C}(I_3)$}
& \multicolumn{2}{@{}c@{}}{$\mathcal{C}(I_3 + \rho)$} \\
\hline
1st & Airliner:       & 99.7\% & aircraft\_carrer: & 94.9\% \\
2nd & Wing:           &  2.6\% & airliner:         &  3.0 \% \\
3rd & Warplane:       &  0.0\% & warplane:         &  1.4 \% \\
4th & Space\_shuttle: &  0.0\% & wing:             &  0.2\% \\
5th & Airship:        &  0.0\% & airship:          & 0.1\% \\
\hline
\end{tabular}
 \end{minipage}
\\
 ~\\~\\
 \begin{minipage}[b]{0.47\textwidth}
 \centering
  (b) $I_4$ \\
\begin{tabular}[t]{@{}l@{~}|@{~}l@{~}r@{~}|@{~}l@{~}r@{}}
\hline
  \multicolumn{1}{@{}c@{~}|@{~}}{Rank}
& \multicolumn{4}{@{}c@{}}{Recognition results and confidence} 
\\   \cline{2-5}     
  \multicolumn{1}{@{}c@{~}|@{~}}{}
& \multicolumn{2}{@{}c@{~}|@{~}}{$\mathcal{C}(I_4)$}
& \multicolumn{2}{@{}c@{}}{$\mathcal{C}(I_4 + \rho)$} \\
\hline
1st & tiger\_cat:                & 81.9\% & Leopard:       & 31.3\% \\
2nd & tabby:                     & 15.8\% & jaguar:        & 10.5\% \\
3rd & Egyptian\_cat:             &  2.0\% & lion:          &  9.3\% \\
4th & Lynx:                      &  0.2\% & snow\_leopard: &  9.2\% \\
5th & Lens\_cap:                 &  0.0\% & cheetah:       &  9.1\% \\
\hline
\end{tabular}
 \end{minipage}
\\
~\\~\\
 \begin{minipage}[b]{0.47\textwidth}
 \centering
  (c) $I_5$ \\
\begin{tabular}[t]{@{}l@{~}|@{~}l@{~}r@{~}|@{~}l@{~}r@{}}
\hline
  \multicolumn{1}{@{}c@{~}|@{~}}{Rank}
& \multicolumn{4}{@{}c@{}}{Recognition results and confidence} 
\\   \cline{2-5}     
  \multicolumn{1}{@{}c@{~}|@{~}}{}
& \multicolumn{2}{@{}c@{~}|@{~}}{$\mathcal{C}(I_5)$}
& \multicolumn{2}{@{}c@{}}{$\mathcal{C}(I_5 + \rho)$} \\
\hline
1st & electric\_guitar: & 96.7\% & Eft:                        & 19.7\% \\
2nd & acoustic\_guitar: &  2.7\% & Banded\_gecko:              & 11.3\% \\
3rd & Violin:           &  0.3\% & European\_fire\_salamander: & 10.2\% \\
4th & Banjo:            &  0.1\% & Common\_newt:               & 10.3\% \\
5th & chello:           &  0.0\% & alligator\_lizard:          &  9.7\% \\
\hline
\end{tabular}
 \end{minipage}
\\
~\\~\\
%
 \begin{minipage}[b]{0.47\textwidth}
 \centering
  (d) $I_6$ \\
\begin{tabular}[t]{@{}l@{~}|@{~}l@{~}r@{~}|@{~}l@{~}r@{}}
\hline
  \multicolumn{1}{@{}c@{~}|@{~}}{Rank}
& \multicolumn{4}{@{}c@{}}{Recognition results and confidence} 
\\   \cline{2-5}     
  \multicolumn{1}{@{}c@{~}|@{~}}{}
& \multicolumn{2}{@{}c@{~}|@{~}}{$\mathcal{C}(I_6)$}
& \multicolumn{2}{@{}c@{}}{$\mathcal{C}(I_6 + \rho)$} \\
\hline
1st & Plastic\_bag:    & 96.2\% & sock:         & 22.5\% \\
2nd & brassiere:       &  1.0\% & brassiere:    &  8.6\% \\
3rd & Toilet\_tissue:  &  0.2\% & pillow:       &  7.8\% \\
4th & diaper:          &  0.2\% & diaper:       &  7.8\% \\
5th & sulphur-crested\_cockatoo: &  0.2\% & handkerchief: &  7.8\% \\
\hline
\end{tabular}
 \end{minipage}
\\
~\\~\\
 \begin{minipage}[b]{0.47\textwidth}
 \centering
  (e) $I_7$ \\
\begin{tabular}[t]{@{}l@{~}|@{~}l@{~}r@{~}|@{~}l@{~}r@{}}
\hline
  \multicolumn{1}{@{}c@{~}|@{~}}{Rank}
& \multicolumn{4}{@{}c@{}}{Recognition results and confidence} 
\\   \cline{2-5}     
  \multicolumn{1}{@{}c@{~}|@{~}}{}
& \multicolumn{2}{@{}c@{~}|@{~}}{$\mathcal{C}(I_7)$}
& \multicolumn{2}{@{}c@{}}{$\mathcal{C}(I_7 + \rho)$} \\
\hline
1st & Promontory:     & 96.6\% & alp:              & 17.8\% \\
2nd & seashore:       &  1.7\% & Irish\_wolfhound: &  8.8\% \\
3rd & cliff:  :       &  1.4\% & marmot:           &  7.5\% \\
4th & bacon:          &  0.2\% & timber\_wolf:     &  7.4\% \\
5th & lakeside:       &  0.0\% & bighorn:          &  7.4\% \\
\hline
\end{tabular}
 \end{minipage}

\end{table}

\subsection{Experiment 4: Other examples by DCT-based methods}

In the final experiment, we attempted to generate AEs using the proposed
DCT-based method for other images of ImageNet-1000 under the accuracy
versus perturbation amount scenario.
In this experiment, we added a solution candidate $\Vec{x}_{\it 0}$
whose all variables were set to 0 into an initial population.
Other experimental conditions were the same as those in experiment 3.
Fig.~\ref{fig:rst_ae_image_dct}(a) shows target original images whose resolution 
was changed to $224 \times 224$.
As in experiment 3, class labels similar to the original ones were
regarded as correct ones, as shown in Table~\ref{tbl:class_labels}.

Fig.~\ref{fig:rst_ae_image_dct}(b) shows the distributions of obtained
non-dominated solutions.
Adding $\Vec{x}_{\it 0}$ allowed the proposed method to clarify the
trade-off relationship between the accuracy and the perturbation amount.
Note that RMSE of $\Vec{x}_{\it 0}$ was not zero because of the effect
of DCT and inverse DCT process.

Fig.\ref{fig:rst_ae_image_dct}~(c) and (d) shows examples of 
generated AEs and their perturbation patterns,
respectively.
Different types of perturbed patterns could be seen; AEs for $I_4$ and
$I_5$ include small numbers of bright pixels, whereas AEs for $I_3$,
$I_6$, and $I_7$ include striped and thin striped patterns.
This demonstrates that the proposed method could adaptively generate AEs
according to the target clean image properties.

Table~\ref{tbl:rst_ae_image_dct} shows the recognized classes and
corresponding confidence scores. 
Here we focus on the results in each image.
Because $I_3$ is an image of the front part of an airplane which
involves less textures, there are very few classes that can induce
misrecognition, resulting in erroneous recognition on label
'aircraft\_carrier'.
Other images $I_4$ through $I_7$ were the objects involving high
frequency components and characteristic colors compared to $I_3$ and
$I_6$, then their AEs made the classifier misclassified to various
classes.
Interestingly, $I_5$ and $I_7$ were erroneously recognized as various
animals, whereas $I_6$ was misclassified mainly as artificial things.

\del{
ImageNet-1000のその他の画像を対象として，DCT-based methodを用いてAEの生
成を試みた． 
$f_1$を正答率，$f_2$を$\Vec{\rho}$の$l_2$ノルムとし，2目的の問題として定式
化した．
$N_P = 10$とし，設計変数は1,424次元となった．
集団サイズを500，世代数の上限を1,000とした．
対象モデルを学習済のVGG16とし，正しいクラス$C(I)$と類似するクラス集合
$\mathcal{N}(C(I))$以外に誤認識させるようにした．\onote{【数式で表現】}
初期集団の中に，全ての変数の値が0，すなわち$I$に相当する個体を1個，含ま
せることとした．

実験に用いた5枚の画像を図\ref{fig:rst_ae_image_dct}(a)に示す．

得られた非劣解集合を図\ref{fig:rst_ae_image_dct}(b)に示す．
全ての変数の値が0の解候補であっても，DCTと逆DCTを経由した後に$f_2$の値を
算出するため，$f_2$の値が0にならない点に留意されたい．

}

\del{
\onote{【20190118の結果】}
\begin{verbatim}
実験1, 実験2:
 - f1: accuracy
 - f2: RMSE
% - var: 1424D DCT
 - 初期個体に入力画像そのものを入れた（全ての変数を0）
 - 飛行機（airliner）
   - 飛行船（air ship），スペースシャトル（space shuttle），軍用戦闘機（war...），
     羽根（wing），飛行機（plane）もう一種類
   - accuracy: 99\% ->
   - 1800世代
     - 1000世代も，少しずつRMSEが減っている

 - 10パタン＋何もいじらないパターン
 - 世代数を1000に変更，個体数500
 - plastic bag
   - 寝袋，バッグ，は除外
 - RMSE: 0.005
   - 正しい認識難しい，妨害されやすい -> パレート解少ない

 - AE: 斜め線
	
\end{verbatim}
}

\subsection{Discussion}

Although some of the above experiments involve high dimensional problems
whose number of design variable exceeds 1,000, the proposed method could
successfully generated AEs under black-box condition, i.e., without
gradient information and other internal information of target
classifiers except final recognition result (a class label and its
confidence score).
The above results revealed that the potential of EMO to AE design,
though there is no guarantee that the obtained solutions were globally
optima.
It is possible that an AE design problem involves a highly multimodal
fitness landscape including many promising quasi-optimal solutions, 
which EMO is appropriate for finding.

\del{
本研究では，AEの生成を設計変数の総数が1000を超える大規模な多目的最適化問
題として定式化し，ロバストなAEなどの設計を試みた．
得られた解が大域的に最適である保証はないものの，十分有用な準最適解を得ら
れていることから，EMOのポテンシャルを示したといえる．
おそらくこれは，対象問題が多峰性の適応度景観をなしており，EMOによって一
部の準最適解を発見できたためと考える．
\onote{【変数間の依存性については？確認した方がいい？】}
局所探索●●とのハイブリッドなどによる探索の効率化や，より多様性を強化す
る工夫等は今後の重要な課題である．
}

\section{Conclusions}

This paper proposes an evolutionary multi-objective optimization approach
to design adversarial examples that cannot be correctly recognized by
machine learning models.
The proposed method is black-box method that does not require internal
information in the target models, and produces various AEs by
simultaneously optimizing multiple objective functions that have
trade-off relationship.
Experimental resultse showed the potentials of the proposed EMO-based
approach;
e.g., the proposed method could produce various AEs that have different
properties from ones generated by the previous EC- and gradient-based
methods,
and AEs robust against image rotation.
This paper also demonstrated that the DCT-based method could generate
AEs for higher resolution images.

On the other hand, the proposed method has many rooms for
improvement from the viewpoint of comprehensively generating more
diverse solutions.
Introducing schemes to promote search exploration and to reduce problem
dimension, and hybridization with local search are our important future
work.
The flexibility of EMO for designing objective functions would allow
emerging new techniques to design AEs.

\del{
本質的にtrade offな関係にある複数の目的関数を含むAE生成問題に対して，EMO
によって多様なAEを包括的に生成する方式を提案した．
画素数と輝度値変更幅とのトレードオフを考慮したAEの生成や，画像変換に対し
て頑健なAEの生成をモデル化できることを示した．

本実験で示した結果は，設計変数の多さに対して個体数等が十分でないため実験
結果に関しては改善の余地が多々ある．
条件を満たすAEは複数存在することから，好適なAEを生成できたものの，
今後，ローカルサーチの併用，設計変数の削減，アルゴリズムの改良等，解法の
改善を行うことは重要な課題である．
}



\end{document}